\documentclass[letterpaper,journal]{IEEEtran}
\usepackage{amsmath,amsfonts,amssymb}
\usepackage{algorithmic}
\usepackage{algorithm}
\usepackage{array}
\usepackage{tabularx}
\usepackage{booktabs}
\usepackage{multirow}
\usepackage[table]{xcolor}
\usepackage{colortbl}
\usepackage{placeins}
\usepackage{arydshln}
\usepackage[colorlinks=true,linkcolor=blue,citecolor=blue,urlcolor=blue]{hyperref}
\usepackage{longtable}
\usepackage[caption=false,font=normalsize,labelfont=sf,textfont=sf]{subfig}
\usepackage{textcomp}
\usepackage{stfloats}
\usepackage{url}
\usepackage{verbatim}
\usepackage{graphicx}
\usepackage{cite}
\usepackage{orcidlink}
\usepackage{rotating}
\usepackage{balance}
\def\BibTeX{{\rm B\kern-.05em{\sc i\kern-.025em b}\kern-.08em
    T\kern-.1667em\lower.7ex\hbox{E}\kern-.125emX}}

\definecolor{bestcolor}{RGB}{248,215,218}
\definecolor{secondcolor}{RGB}{219,233,251}
\definecolor{avgcolor}{RGB}{226,239,218}
\definecolor{groupgray}{RGB}{235,235,235}
\definecolor{groupcolor}{RGB}{242,242,242}
\newcommand{\best}[1]{#1}
\newcommand{\second}[1]{#1}
\newcommand{\blockheader}[1]{\multicolumn{1}{l}{#1}\\}
\newcommand{\seriesheader}[1]{\multicolumn{1}{l}{#1}\\}
\newcommand{\groupsep}{}
\newcommand{\orcidYuqiao}{0009-0006-9937-1232}
\newcommand{\orcidFei}{0009-0004-1142-6434}
\newcommand{\orcidKun}{0000-0001-5083-2145}
\newcommand{\orcidYanyan}{0000-0001-8818-6740}

\title{RRS-10K: A Multitask Vision-Language Model Benchmark for Rare Remote Sensing Image Interpretation}

\author{Yuqiao Lai\,\orcidlink{\orcidYuqiao}
Jiancheng Qi,
Fei Wang\,\orcidlink{\orcidFei},
Yuxin Liu,\\
Kun Li\,\orcidlink{\orcidKun},~\IEEEmembership{Member,~IEEE},
Ye Chen,
Yan Gao,
and Yanyan Wei\,\orcidlink{\orcidYanyan}%
\thanks{This work was supported in part by the National Natural Science Foundation of China under Grant 62272144, in part by the National Key Research and Development Program of China under Grant 2024YFB3311602, and in part by the Independent Innovation Science Fund of National University of Defense Technology under Grant 25-ZZCX-ZX-RWSK-25.}%
\thanks{Yuqiao Lai and Jiancheng Qi contributed equally to this work.}%
\thanks{Yuqiao Lai, Jiancheng Qi, Ye Chen, and Yan Gao are with the Laboratory of Intelligent Language Processing, National University of Defense Technology, Nanjing 210039, China.}%
\thanks{Fei Wang and Yanyan Wei are with Hefei University of Technology, Hefei 230009, China.}%
\thanks{Fei Wang and Yuxin Liu are also with the Institute of Artificial Intelligence, Hefei Comprehensive National Science Center, Hefei 230026, China.}%
\thanks{Kun Li is with United Arab Emirates University, Al Ain, P.O. Box 15551, United Arab Emirates.}%
\thanks{Corresponding authors: Yan Gao (e-mail: gaoyan22@nudt.edu.cn) and Yanyan Wei (e-mail: weiyy@hfut.edu.cn).}%
\thanks{Code, dataset, and models will be available online at https://github.com/weiweinuonuox/RRS-10K-A-Benchmark-for-rare-remote-sensing-image-interpretation.}%
\thanks{Manuscript submitted to IEEE Journal of Selected Topics in Applied Earth Observations and Remote Sensing on May 18, 2026.}%
}

\begin{document}

\maketitle

\begin{abstract}
Vision-language models (VLMs) have achieved strong performance on general remote sensing tasks. However, their capability for rare scenes remains insufficiently understood, because existing benchmarks are dominated by common urban and rural imagery. To address this gap, we present RRS-10K, a benchmark for rare remote sensing image interpretation. RRS-10K contains 10,738 military-related remote sensing images and corresponding multiple format question-answer pairs for comprehensive evaluation. All of the images are collected from first-hand sources and organized into three capability dimensions, six sub-dimensions, and 20 leaf tasks, covering perception, reasoning, and robustness. To improve the quality of multiple-choice questions, we introduce a similarity-based distractor filtering strategy (SDFS) during benchmark construction. We further evaluate 52 representative models and show that current VLMs achieve only moderate zero-shot performance on rare remote sensing image interpretation, with clear weaknesses in visual grounding, referring segmentation, and complex semantic reasoning tasks. RRS-10K enables systematic analysis of failure modes in long-tail remote sensing interpretation and provides guidance for developing more reliable remote sensing VLMs.
\end{abstract}


\section{Introduction}
\label{sec:introduction}


\IEEEPARstart{R}{are} remote sensing image interpretation is of considerable importance for downstream earth observation applications. It requires a timely and reliable understanding of uncommon yet high-value scenes, such as disaster assessment~\cite{ref-1}, marine surveillance~\cite{ref-2}, and national security~\cite{ref-3}. Unlike general remote sensing scenes dominated by urban or agricultural land covers, rare scenes are characterized by semantic ambiguity, complex background, and long-tail category distribution, making them more challenging to interpret~\cite{ref-4}. Vision-language models (VLMs), with their strong cross-modal semantic alignment, have achieved impressive performance on general remote sensing tasks, such as scene classification~\cite{ref-5,ref-6}, object detection~\cite{ref-7,ref-8}, and segmentation tasks~\cite{ref-9,ref-10}. However, the model's ability to interpret rare scenes remains insufficiently investigated, motivating the construction of a dedicated benchmark for such scenes.


VLMs have created a new paradigm for image understanding by bridging visual and linguistic modalities. CLIP~\cite{ref-11} pioneered contrastive vision-language pretraining, demonstrating remarkable zero-shot capabilities. LLaVA~\cite{ref-12} connected visual encoders with large language models through visual instruction tuning. More recently, generative VLMs such as GPT-4o and Qwen3-VL have achieved state-of-the-art performance on general benchmarks. However, their direct application to remote sensing faces significant challenges due to domain gaps. Motivated by this challenge, researchers have developed remote sensing vision-language models (RS-VLMs). RS-CLIP~\cite{ref-13} first introduced VLMs into the remote sensing domain, preliminarily validating their feasibility for zero-shot remote sensing image classification. RemoteCLIP~\cite{ref-14} subsequently conducted large-scale remote sensing image-text pretraining from scratch. Beyond contrastive learning approaches, instruction-tuned RS-VLMs have emerged to support more sophisticated reasoning. EarthGPT~\cite{ref-15} and SkyEyeGPT~\cite{ref-16} further advanced toward unified end-to-end multimodal intelligence for comprehensive remote sensing image understanding. 

As RS-VLMs continue to advance, the need for systematic benchmarks has become increasingly evident. Early benchmark construction for RS-VLMs concentrated on specific tasks, such as scene classification, visual question answering (VQA), and captioning. For example, RSVQA~\cite{ref-17} established remote sensing visual question answering (RSVQA) as a benchmark task, while EarthVQA~\cite{ref-18} further emphasized relational reasoning, counting, and geo-spatial analysis. Beyond VQA, VRSBench~\cite{ref-19} broadened evaluation to multiple vision-language tasks, including captioning and visual grounding. Recent efforts have shifted toward the construction of systematic benchmarks. RSVLM-QA~\cite{ref-20} enriched the benchmark with a hierarchical structure. Building on this trend, studies have moved beyond broad task coverage to examine whether RS-VLMs can support complex reasoning tasks and remain reliable under more realistic Earth observation conditions: VLRS-Bench~\cite{ref-21} focuses specifically on complex remote sensing reasoning, including cognition, decision-making, and prediction, while OmniEarth~\cite{ref-22} is constructed under realistic Earth observation scenarios with multi-source data and diverse output forms. Collectively, benchmarks for RS-VLMs are evolving from task-oriented to hierarchical.

Recent benchmarks have substantially expanded evaluation in terms of task breadth, capability hierarchy, and answer formats. However, existing benchmarks are largely built upon general scenes, where urban and rural scenes dominate data distribution. Compared with common scenes, rare scenes are often characterized by sparse and infrequent targets, limited annotated resources, stronger dependence on domain-specific knowledge, and long-tail category distribution~\cite{ref-4,ref-23}. Meanwhile, rare-scene resources remain highly fragmented, being largely confined to task-specific datasets for military aircraft~\cite{ref-24}, ships~\cite{ref-25}, or vehicles~\cite{ref-26}. Therefore, existing benchmarks remain insufficient for rare-scene settings, as they cannot fully capture the unique challenges posed by rare target distributions, scarce annotations, and complex semantic reasoning required in such scenes, as summarized in Table~\ref{tab1}.

\newcommand{\rrsshortvline}{\vrule width \arrayrulewidth height 1.5ex depth 0.35ex}
\newcolumntype{C}[1]{>{\centering\arraybackslash}m{#1}}

\begin{table*}[!t]
\caption{Comparison of RRS-10K With Different Benchmarks for Remote Sensing Scenes\label{tab1}}
\centering
\begingroup
\setlength{\tabcolsep}{0pt}
\begin{tabular}{@{}C{1.10in}|C{0.52in}|C{0.55in}|C{0.55in}|C{0.55in}|C{0.42in}|C{0.68in}|C{0.72in}|C{1.95in}@{}}
\hline
\multicolumn{1}{@{}C{1.10in}!{\rrsshortvline}}{Dataset} &
\multicolumn{1}{C{0.52in}!{\rrsshortvline}}{Images} &
\multicolumn{1}{C{0.55in}!{\rrsshortvline}}{Aircraft} &
\multicolumn{1}{C{0.55in}!{\rrsshortvline}}{Vehicle} &
\multicolumn{1}{C{0.55in}!{\rrsshortvline}}{Vessels} &
\multicolumn{1}{C{0.42in}!{\rrsshortvline}}{Sites} &
\multicolumn{1}{C{0.68in}!{\rrsshortvline}}{Reasoning} &
\multicolumn{1}{C{0.72in}!{\rrsshortvline}}{Rare Scenes} &
\multicolumn{1}{C{1.95in}@{}}{Answer Type} \\
\hline
\noalign{\vskip 1.0pt}
MAR20~\cite{ref-24}        & 3,842  & $\checkmark$ & $\times$     & $\times$     & $\times$     & $\times$     & $\checkmark$ & BBox \\
MVRSD~\cite{ref-26}        & 3,000  & $\times$     & $\checkmark$ & $\times$     & $\times$     & $\times$     & $\checkmark$ & BBox \\
MSTAR~\cite{ref-27}        & 1,500  & $\times$     & $\checkmark$ & $\times$     & $\times$     & $\times$     & $\checkmark$ & BBox \\
NWPUVHR-10~\cite{ref-28}   & 800    & $\checkmark$ & $\checkmark$ & $\times$     & $\times$     & $\times$     & $\checkmark$ & BBox \\
HRSID~\cite{ref-25}        & 5,604  & $\times$     & $\times$     & $\checkmark$ & $\times$     & $\times$     & $\checkmark$ & BBox, Mask \\
DOTA~\cite{ref-29}         & 2,806  & $\checkmark$ & $\checkmark$ & $\checkmark$ & $\checkmark$ & $\times$     & $\times$     & BBox \\
SAM-Data~\cite{ref-30}     & 300    & $\times$     & $\times$     & $\times$     & $\checkmark$ & $\checkmark$ & $\checkmark$ & Free Form \\
VLRS-Bench~\cite{ref-21}   & 2,000  & $\checkmark$ & $\checkmark$ & $\checkmark$ & $\checkmark$ & $\times$     & $\times$     & MCQ, BBox, Free Form, Mask \\
CHOICE~\cite{ref-31}       & 10,507 & $\checkmark$ & $\checkmark$ & $\checkmark$ & $\checkmark$ & $\checkmark$ & $\times$     & MCQ, BBox, Free Form, Mask \\
\noalign{\vskip 1.0pt}
\hline
\multicolumn{1}{@{}C{1.10in}!{\rrsshortvline}}{\textbf{RRS-10K}} &
\multicolumn{1}{C{0.52in}!{\rrsshortvline}}{\textbf{10,738}} &
\multicolumn{1}{C{0.55in}!{\rrsshortvline}}{$\checkmark$} &
\multicolumn{1}{C{0.55in}!{\rrsshortvline}}{$\checkmark$} &
\multicolumn{1}{C{0.55in}!{\rrsshortvline}}{$\checkmark$} &
\multicolumn{1}{C{0.42in}!{\rrsshortvline}}{$\checkmark$} &
\multicolumn{1}{C{0.68in}!{\rrsshortvline}}{$\checkmark$} &
\multicolumn{1}{C{0.72in}!{\rrsshortvline}}{$\checkmark$} &
\multicolumn{1}{C{1.95in}@{}}{MCQ, BBox, Free Form, Mask} \\
\hline
\end{tabular}
\endgroup

\vspace{1mm}
\begin{minipage}{\textwidth}
BBox: bounding box; MCQ: multiple-choice question; Free Form: open-ended textual answer; Mask: pixel-level segmentation mask.
\end{minipage}
\end{table*}

\begin{figure}[!t]
    \centering
    \includegraphics[width=\columnwidth]{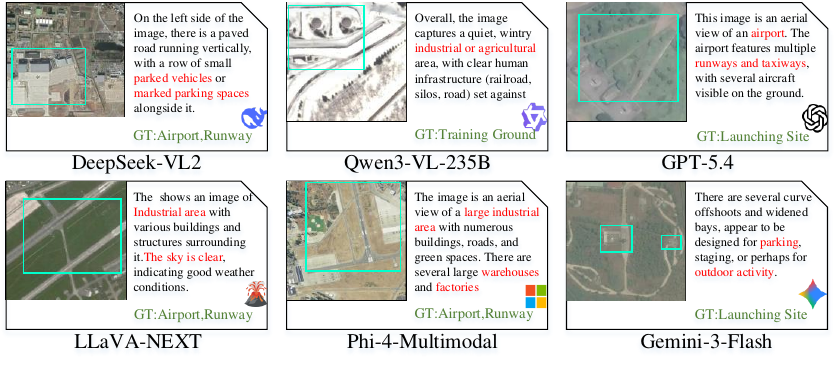}
    \caption{Performance of general VLMs on rare-scene remote sensing images. These models frequently produce ambiguous or incorrect interpretations in military domains, especially when distinguishing functionally similar scenes such as airports, training grounds, and launching sites.}
    \label{fig1}
\end{figure}

\begin{itemize}
    \item \textbf{Insufficient Rare-scene Data:} Rare-scene remote sensing datasets remain scarce, difficult to access at scale, and expensive to annotate reliably. Most existing resources are proprietary, such as Maxar SecureWatch~\cite{ref-2}, or only partially open-source, such as MSTAR~\cite{ref-27}. Even publicly accessible datasets are limited in annotation scale; for instance, SAMData~\cite{ref-30} contains only 300 manually annotated images. In addition, some benchmarks are repurposed from public captioning or VQA datasets, which may overlap with the pretraining corpora of VLMs ~\cite{ref-20,ref-32}, thereby introducing potential data leakage and compromising evaluation objectivity. Consequently, current models often exhibit unsatisfactory performance in sparse domains such as military scenes, where subtle visual cues and strong domain-specific semantics are required for accurate interpretation, as illustrated in Fig.~\ref{fig1}.
    
    \item \textbf{Fragmented Capability Coverage:} Existing rare-scene datasets are largely organized upon isolated tasks, such as military target detection~\cite{ref-24,ref-25,ref-26} and military-oriented image captioning~\cite{ref-3,ref-33,ref-34}. Most of these datasets are limited to image-level understanding or object-level recognition. Therefore cannot systematically evaluate the capabilities of RS-VLMs under rare scenes, including fine-grained perception, multi-step reasoning, temporal understanding, and robustness under complex conditions.
    
    \item \textbf{Lack of Comprehensive Evaluation:} Rare-scene understanding requires not only fundamental capabilities, such as scene recognition and object identification, but also higher-level reasoning capabilities over spatial relations, scene functionality, and temporal dynamics. However, existing benchmarks~\cite{ref-17,ref-20,ref-35} mainly focus on shallow visual cues, simple counting, and coarse positional judgment. Such evaluations are insufficient to expose the intrinsic capability bottlenecks of current models and provide limited guidance for future development.
\end{itemize}

To address these challenges and evaluate the capabilities of current VLMs, we propose RRS-10K, a multitask benchmark for rare remote sensing image interpretation. The contributions of this paper are as follows:

\begin{itemize}
\item We introduce RRS-10K, a multitask benchmark for rare remote sensing image interpretation that contains 10,738 images collected from 24 countries and 57 cities. Built from first-hand remote sensing imagery of military scenes, RRS-10K is approximately 35 times larger than that of SAM-Data~\cite{ref-30}, the most closely related benchmark considered. RRS-10K supports multiple evaluation formats, including multiple-choice questions (MCQs), open-ended question answering, and pixel-level segmentation. To ensure benchmark quality, a human-centered hybrid annotation pipeline together with a Similarity-based Distractor Filtering Strategy (SDFS) is employed.

\item We establish a hierarchical task taxonomy for rare remote sensing evaluation, consisting of three capability dimensions, six sub-dimensions, and 20 leaf tasks. Specifically, capability dimensions capture core evaluation objectives of Perception, Reasoning, and Robustness. Each objective is further decomposed into sub-dimensions, and leaf tasks operationalize them as concrete downstream tasks. This design enables a unified assessment of rare-scene understanding beyond fragmented task-specific evaluation.

\item We conduct an extensive evaluation of 52 representative models on RRS-10K, including 43 VLMs and 9 referring segmentation models, providing a systematic analysis of their performance. This evaluation establishes a unified empirical basis for diagnosing model behavior across perception,  reasoning, and robustness, offering actionable insights for future development of RS-VLMs.
\end{itemize}

\section{Related Work}
\label{sec:related}

\subsection{Rare Remote Sensing Image Interpretation}

Rare remote sensing image interpretation refers to the understanding of low-frequency, domain-specific remote sensing scenarios with strong application specificity. In this study, representative rare-scene cases mainly include military scenes~\cite{ref-24,ref-26}, disaster assessment~\cite{ref-1}, and maritime surveillance~\cite{ref-25}. Compared with general remote sensing settings, such scenes are more likely to suffer from limited data availability, high annotation difficulty, and stronger dependence on professional knowledge~\cite{ref-4,ref-23}. These characteristics make rare-scene interpretation a challenging problem.

Early studies in this direction were task-specific, focusing on target detection, recognition, and segmentation: MAR20~\cite{ref-24} establishes a fine-grained benchmark for military aircraft recognition, while HRSID~\cite{ref-25} focuses on ship detection and instance-level analysis in high-resolution harbor imagery. DOTA~\cite{ref-29} provides large-scale oriented object annotations for aircraft, ships, vehicles, and man-made facilities, and SSDD~\cite{ref-36} extends ship detection to the SAR domain. For ground targets, MSTAR~\cite{ref-27} supports SAR-based military vehicle interpretation, whereas MVRSD~\cite{ref-26} provides optical imagery for military vehicle detection across diverse scenes. At the pixel level, HRSID~\cite{ref-25} and NWPU VHR-10~\cite{ref-28} further support fine-grained target delineation and shape-aware analysis through segmentation annotations. Collectively, these datasets have substantially advanced rare-scene perception.

More recent efforts have begun to move beyond conventional recognition pipelines toward language-driven interpretation and multimodal interaction. In the military and defense domain, MiliChat~\cite{ref-37}, MGPT~\cite{ref-38}, and TRACLM~\cite{ref-39} explore large-language-model-based question answering and situational understanding, indicating the potential of foundation models for specialized scene interpretation. In the multimodal setting, SAMChat~\cite{ref-30} further incorporates vision-language reasoning to support rare target understanding. These studies suggest that rare-scene interpretation is gradually evolving from closed-set visual recognition toward more flexible instruction-following and semantic reasoning paradigms.

Nevertheless, progress in rare-scene remote sensing resources remains fragmented. At the resource level, relevant data are still either proprietary and difficult to access at scale, such as Maxar SecureWatch~\cite{ref-2}, or limited in openness and scale, such as MSTAR~\cite{ref-27}. At the task level, most existing datasets are still designed for isolated objectives, including target detection, fine-grained recognition, or instance segmentation, and therefore mainly support local perception rather than holistic rare-scene understanding. Even recent multimodal efforts such as SAMChat~\cite{ref-30} only begin to introduce language-driven reasoning, but remain constrained by very limited data scale and relatively single annotation and task settings. Consequently, current resources still cannot provide a unified evaluation of fine-grained perception, spatial reasoning, temporal interpretation, and robustness within a single framework. As a result, although prior studies have laid an important foundation for rare RS scene interpretation, a systematic benchmark tailored to rare, long-tail, and sparsely distributed targets is still lacking.

\subsection{Benchmarks for RS-VLMs}
With the rapid development of RS-VLMs, corresponding benchmarks have also continued to evolve. Early remote sensing benchmarks mainly focused on image captioning and VQA. UCM-Captions~\cite{ref-40} provided one of the earliest benchmarks for remote sensing image captioning, supporting scene-level language description. RSICD~\cite{ref-32} subsequently extended this line of research by constructing a larger captioning dataset with richer semantic annotations. RSVQA~\cite{ref-17} further moved benchmark development toward question answering and became the first large-scale RS-VQA dataset. On this basis, the research community has developed more comprehensive benchmarks. For example, CHOICE~\cite{ref-31} is the first benchmark specifically designed to evaluate remote sensing VLMs, focusing on understanding and reasoning ability. VRSBench~\cite{ref-19} focuses on understanding spatial relationships and provides fine-grained evaluation dimensions. RSVLM-QA~\cite{ref-20} integrates multiple remote sensing datasets and enriches VQA evaluation with detailed captions, spatial relations, and counting questions. Together, these benchmarks have driven remote sensing VLM evaluation from single-task settings toward integrated multi-dimensional assessment.

While existing benchmarks perform well in supporting RS-VLM evaluation for general remote sensing scenarios, they remain insufficient for rare-scene settings. First, available data for rare remote sensing scenes remain highly restricted: most datasets are proprietary, such as Maxar SecureWatch~\cite{ref-2}, or only partially open-source, such as MSTAR~\cite{ref-27}, while even the few publicly accessible resources, such as SAMData~\cite{ref-30}, contain only a very limited number of manually annotated samples, which constrains annotation diversity and scene coverage. Second, existing rare-scene datasets are still dominated by single-task formulations, such as military target detection~\cite{ref-24}, ship recognition~\cite{ref-25}, and military vehicle detection~\cite{ref-26}, and are largely confined to basic image-level classification or object detection, making them inadequate for evaluating multimodal capabilities required by RS-VLMs, including spatial understanding, semantic reasoning, and robustness. Although recent benchmarks such as GEOBench-VLM~\cite{ref-41} and VRSBench~\cite{ref-19} have expanded evaluation to multiple tasks, they are not specifically designed for rare scenes and therefore cannot fully capture the unique challenges posed by globally rare distributions, rare targets, and long-tail categories. Third, current question sets~\cite{ref-17,ref-20,ref-35} mainly focus on shallow visual semantics, simple counting, and coarse positional judgments, while systematic evaluation of high-level reasoning abilities, especially complex spatial relation understanding, rare-scene contextual inference, and more challenging scenario-oriented reasoning remains largely absent. Finally, the objectivity of existing benchmarks is also affected by potential data leakage, since some of them are adapted from public captioning or VQA datasets that may already have been involved in the training process of VLMs~\cite{ref-20,ref-32}. Such overlap can lead to inflated performance estimates and further compromise evaluation objectivity. Taken together, these limitations suggest that rare-scene RS-VLM evaluation still lacks a dedicated and objective benchmark with sufficient difficulty and broad capability coverage, thereby motivating the construction of RRS-10K.

\subsection{RS-VLMs}

Remote sensing imagery is often characterized by large-scale variations, complex backgrounds, and sparse object distributions. Owing to their multimodal alignment and semantic modeling capabilities, VLMs provide a feasible paradigm for interpreting such scenes. Consequently, researchers have begun to explore RS-VLMs. RS-CLIP~\cite{ref-13} was the first to introduce VLMs into the remote sensing domain and preliminarily verified their feasibility for zero-shot remote sensing tasks. Remote-CLIP~\cite{ref-14} leveraged large-scale remote sensing image-text datasets for large-scale pretraining from scratch, further validating the applicability of VLMs to this domain.

Although CLIP-based RS-VLMs have shown utility in image-text retrieval tasks, they remain limited in handling complex scenarios involving multi-scale, compositional, and language interactions. Accordingly, researchers have started to investigate generative RS-VLMs built upon large language models. RSGPT~\cite{ref-42} first demonstrated the effectiveness of instruction tuning for the remote sensing domain. RSLLaVA~\cite{ref-35} improved the accuracy of fine-grained semantic understanding through optimizing visual feature alignment. GeoChat~\cite{ref-43} exploited the advantages of geospatial semantic information to achieve precise visual localization. LHRS-Bot~\cite{ref-44} focuses on geospatial reasoning tasks and supports fine-grained semantic understanding involving real-world geographic spaces and spatial relationships. On this basis, EarthGPT~\cite{ref-15} and SkyEyeGPT~\cite{ref-16} further advanced toward building unified end-to-end multimodal intelligence, enabling a more comprehensive understanding of remote sensing imagery.

\section{RRS-10K Benchmark}
\label{sec:benchmark}
\subsection{Overview}
\label{sec:overview}

\begin{figure}[!t]
    \centering

    \begin{minipage}[t]{\columnwidth}
        \centering
        \includegraphics[width=\linewidth]{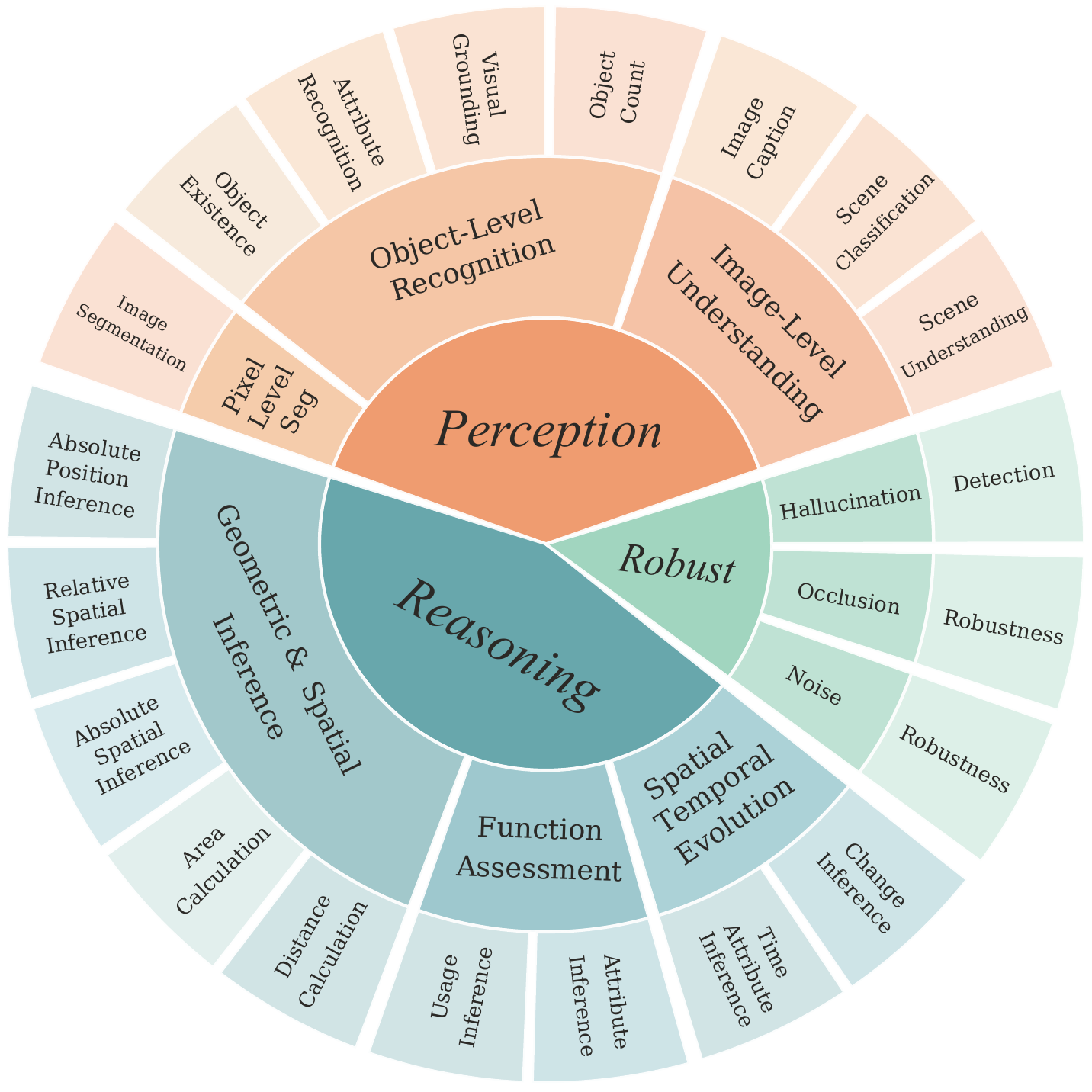}
        
        \vspace{0.1em}
        \small (a) Hierarchical task taxonomy of RRS-10K.
    \end{minipage}

    \vspace{0.6em}

    \newcommand{\subfigh}{0.16\textheight}

\begin{minipage}[t]{0.48\columnwidth}
    \centering
    \includegraphics[width=\linewidth,height=2.45cm]{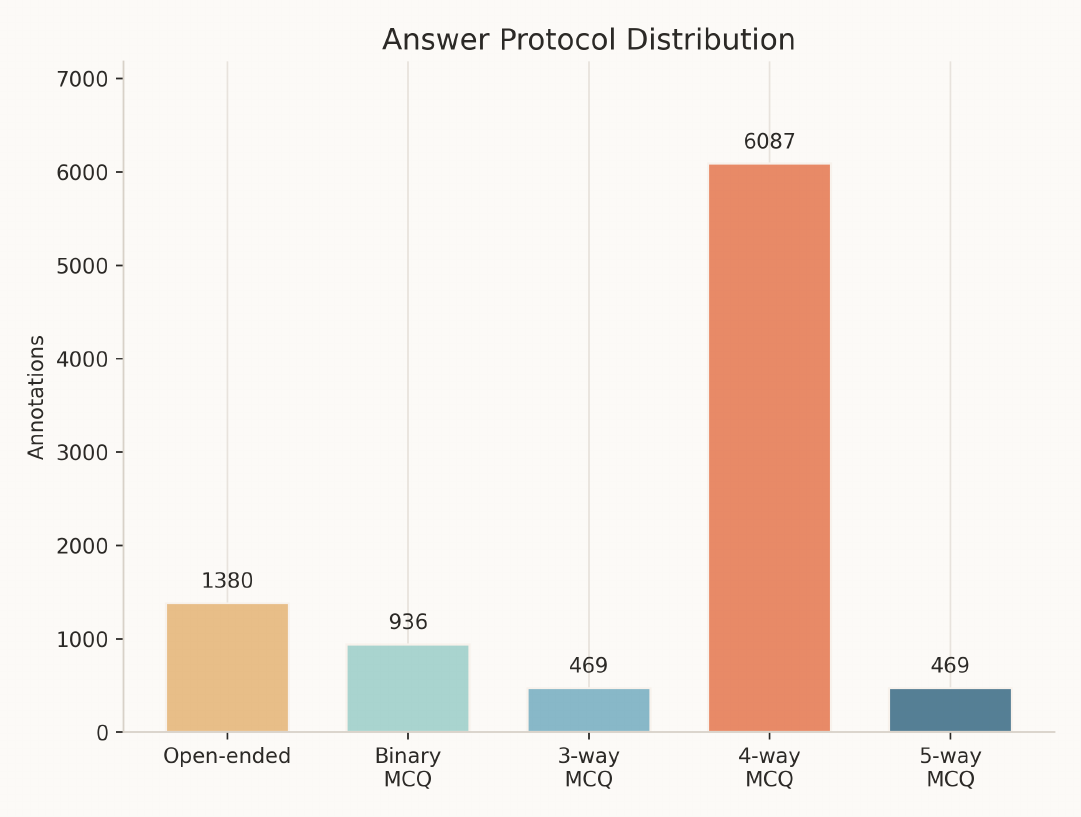}
    
    \small (b) Distribution of answer protocols.
\end{minipage}
\hfill
\begin{minipage}[t]{0.48\columnwidth}
    \centering
    \includegraphics[width=\linewidth,height=2.45cm]{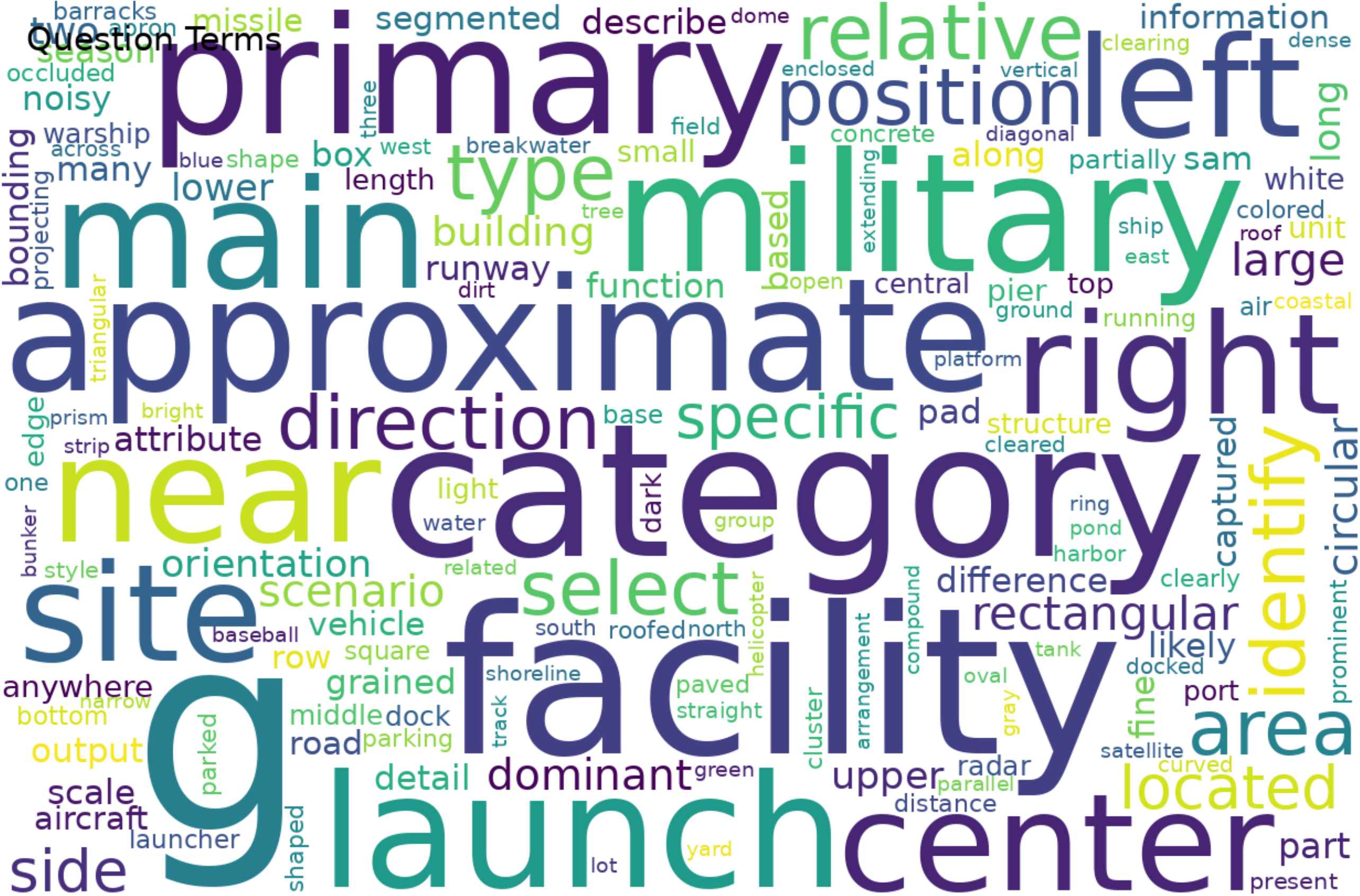}
    
    \small (c) Word cloud of RRS-10K.
\end{minipage}

\begin{minipage}[t]{0.48\columnwidth}
    \centering
    \includegraphics[width=\linewidth,height=2.45cm]{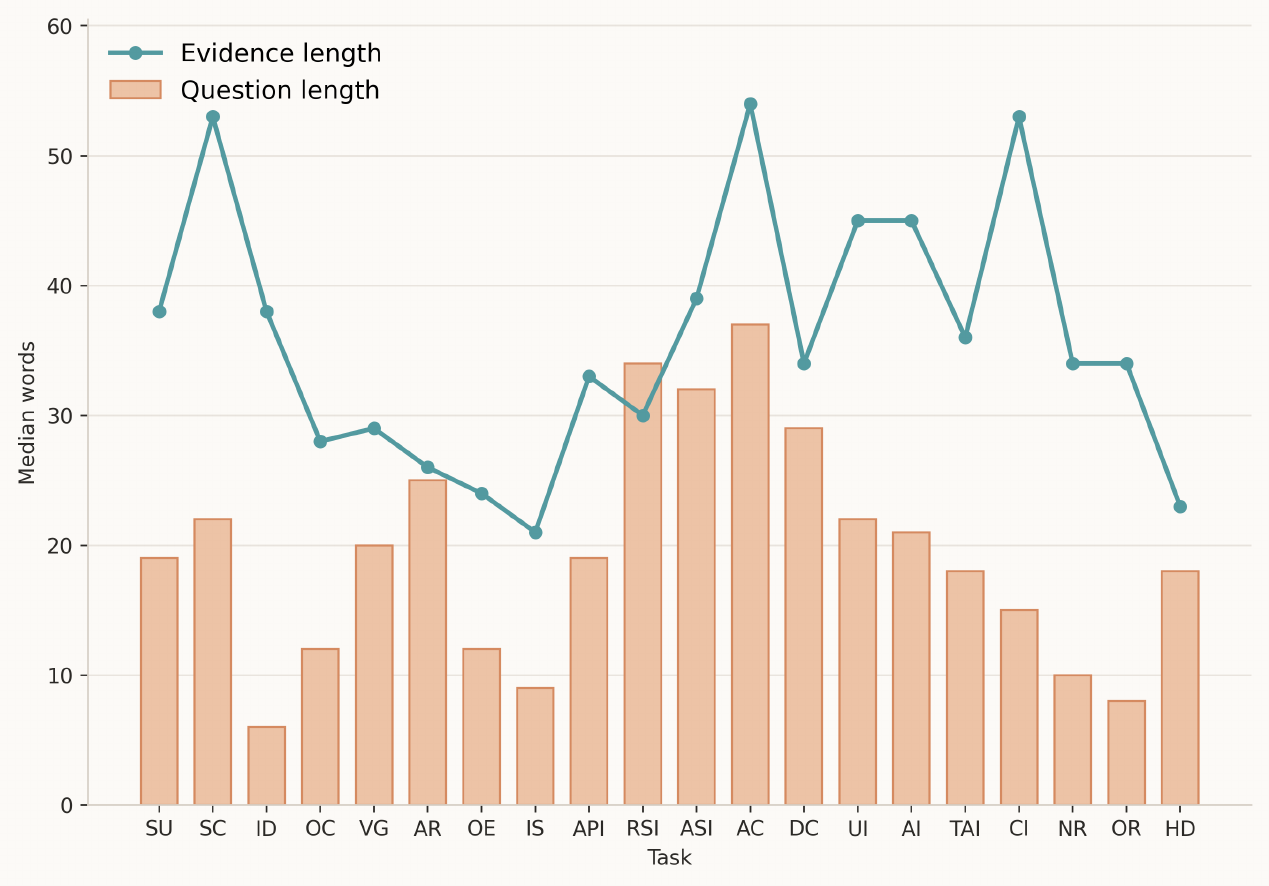}
    
    \small (d) Median lengths of questions and evidence.
\end{minipage}
\hfill
\begin{minipage}[t]{0.48\columnwidth}
    \centering
    \includegraphics[width=\linewidth,height=2.45cm]{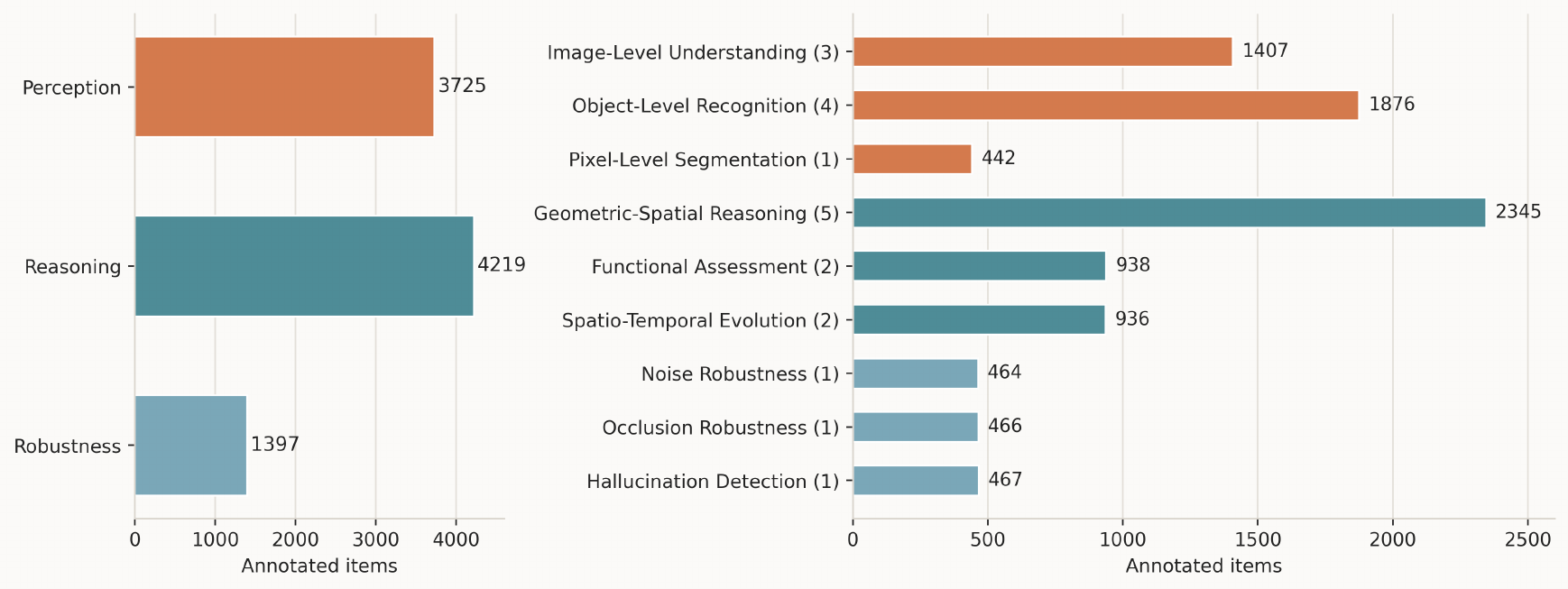}
    
    \small (e) Capability and L2 family coverage.
\end{minipage}

\caption{RRS-10K is organized into a hierarchical taxonomy comprising 3 capability dimensions, 6 sub-dimensions, and 20 leaf tasks, supporting multiple evaluation formats.
(a) Capabilities taxonomy of RRS-10K. 
(b) Distribution of answer protocols, RRS-10K supports multiple evaluation formats, including multiple-choice questions, open-ended responses, bounding boxes, and pixel masks.
(c) Word cloud of question terms, focusing on rare remote sensing targets and facilities. 
(d) Median lengths of questions and evidence annotations.
(e) Capability coverage, showing sample distribution across different evaluation dimensions.}
\label{fig:rrs10k_combined}
\end{figure}

RRS-10K is a multitask and multi-temporal benchmark for rare remote sensing image interpretation, characterized by both diverse answer protocols and a hierarchical task taxonomy. As shown in Fig.~\ref{fig:rrs10k_combined}(a), the benchmark is organized into three capability dimensions, six sub-dimensions, and 20 leaf tasks, covering from global scene understanding to complex semantic target analysis under challenging observation conditions. Fig.~\ref{fig:rrs10k_combined}(b) shows that 4-way MCQs constitute the dominant answer format, while open-ended, binary, 3-way, and 5-way settings provide complementary evaluation scenarios. Fig.~\ref{fig:rrs10k_combined}(c) further indicates that the most frequent question terms are concentrated on facility recognition, spatial localization, directional judgment, and fine-grained category discrimination. Fig.~\ref{fig:rrs10k_combined}(d) shows that questions are generally concise, whereas evidence annotations tend to be longer. Fig.~\ref{fig:rrs10k_combined}(e) summarizes the distribution of samples across capability dimensions and task families, indicating broad coverage of perception, reasoning, and robustness-oriented evaluation, balancing concise question formulation with richer evidence descriptions.

\subsection{Dataset Construction and Quality Control}

\begin{figure*}[!t]
\includegraphics[width=\textwidth]{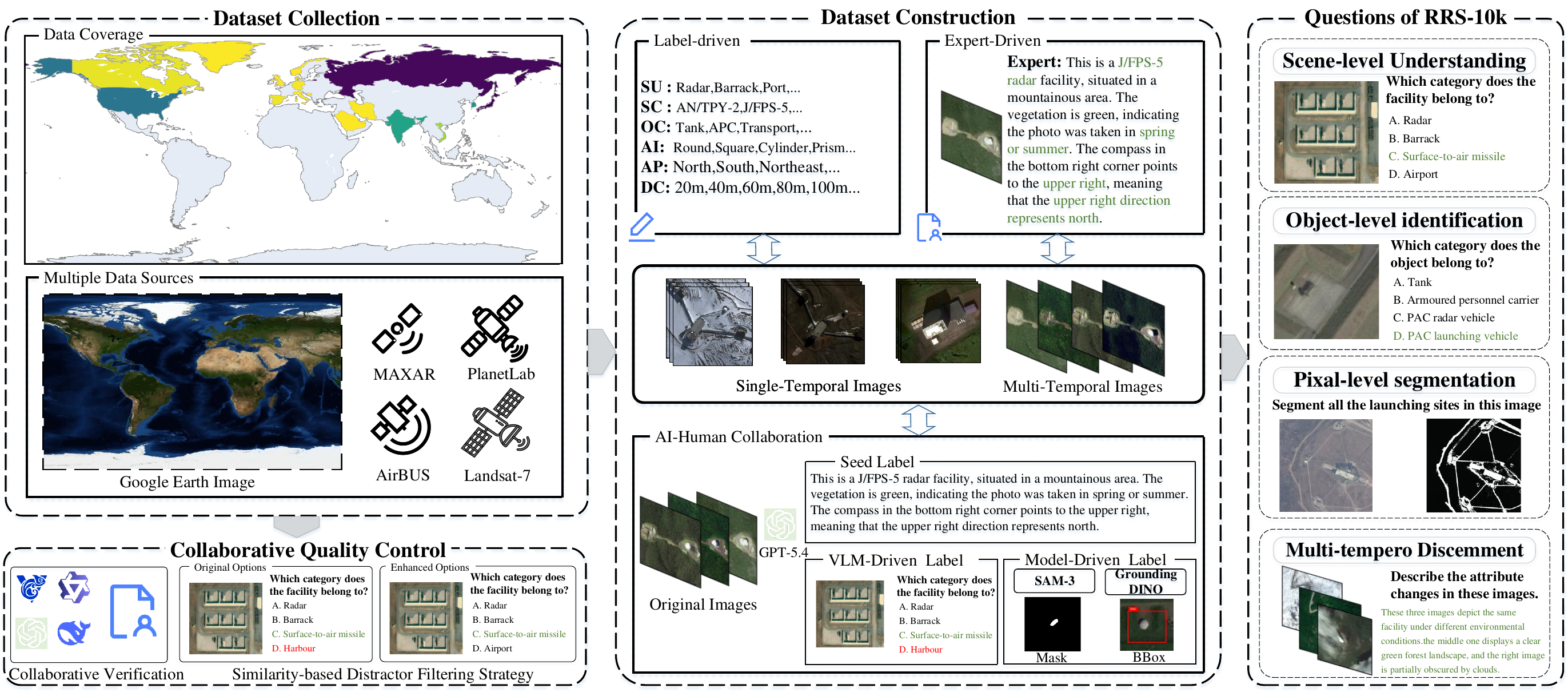}
\caption{Overview of the RRS-10K Benchmark Construction. All remote sensing imagery is sourced from publicly available datasets and undergoes a rigorous quality control process involving both manual verification and VLM-assisted validation.\label{fig3}}
\end{figure*}

Balancing question quality and annotation efficiency is critical during benchmarking, considering the scale of remote sensing scenes. RRS-10K establishes a collaborative pipeline that integrates expert knowledge, open-source information, and model-assisted label, with rigorous quality control maintained throughout the whole process, as illustrated in Fig.~\ref{fig3}.

\subsubsection{Data Acquisition}

To compensate for the limited geographic coverage of existing remote sensing benchmark datasets, we collected high-resolution optical remote sensing imagery from publicly available sources and commercial satellite imagery providers, including Google Earth, Maxar, Airbus, and Planet Labs. RRS-10K covers imagery from 24 countries and regions, with spatial resolutions ranging from sub-meter to decimeter levels. To ensure broad scenario diversity and representative target distribution, this process resulted in a dataset of 10,738 remote sensing images with improved geographic diversity and representative rare-scene coverage.

\subsubsection{Human-Centered Hybrid Annotation Scheme}

During the construction of RRS-10K, the coexistence of semantic annotation and spatial grounding tasks makes the annotation process particularly challenging. To balance dataset scale, annotation efficiency, and professional reliability, we adopt a human-centered hybrid annotation framework that combines expert experience, seed examples, and model-assisted label.

2.1) Incorporation of expert knowledge. Domain experts first provide descriptive prompts for different target categories. These expert-defined descriptions cover key aspects such as target category, structural characteristics, contextual environment, and functional attributes, thereby establishing the primary semantic basis for subsequent annotation.

2.2) Open-source information augmentation. To further enrich the expert knowledge base, we collect relevant textual materials from open-source resources, including \textit{Jane's Defence Weekly}, Wikipedia, and Fandom Military Wiki. These materials serve as complementary knowledge sources and broaden the semantic coverage of the annotations.

2.3) Semi-automated annotation. Based on expert-written prompts and domain-specific textual resources, we employ GPT-5.4 to generate preliminary annotations in the format with image, question, reasoning, and answer. Reasoning component functions not only as an explanation of the final answer, but also as an interpretable intermediate representation that provides structured supervision for vision-language alignment.

2.4) Language-driven foundation model annotation. We develop a language-driven annotation pipeline for visual grounding and referring segmentation. Specifically, based on expert knowledge, GPT-5.4~\cite{ref-45} is used to generate brief descriptions of visually salient targets in each image. These descriptions are then used as prompts for downstream foundation models: Grounding DINO~\cite{ref-46} is employed to generate bounding-box annotations for visual grounding, while SAM-3~\cite{ref-47} is adopted to produce segmentation masks for referring segmentation. All generated annotations are subsequently reviewed and corrected by human annotators to ensure annotation quality.

\subsubsection{Similarity-based Distractor Filtering Strategy}

\begin{figure}[!t]
\includegraphics[width=\columnwidth]{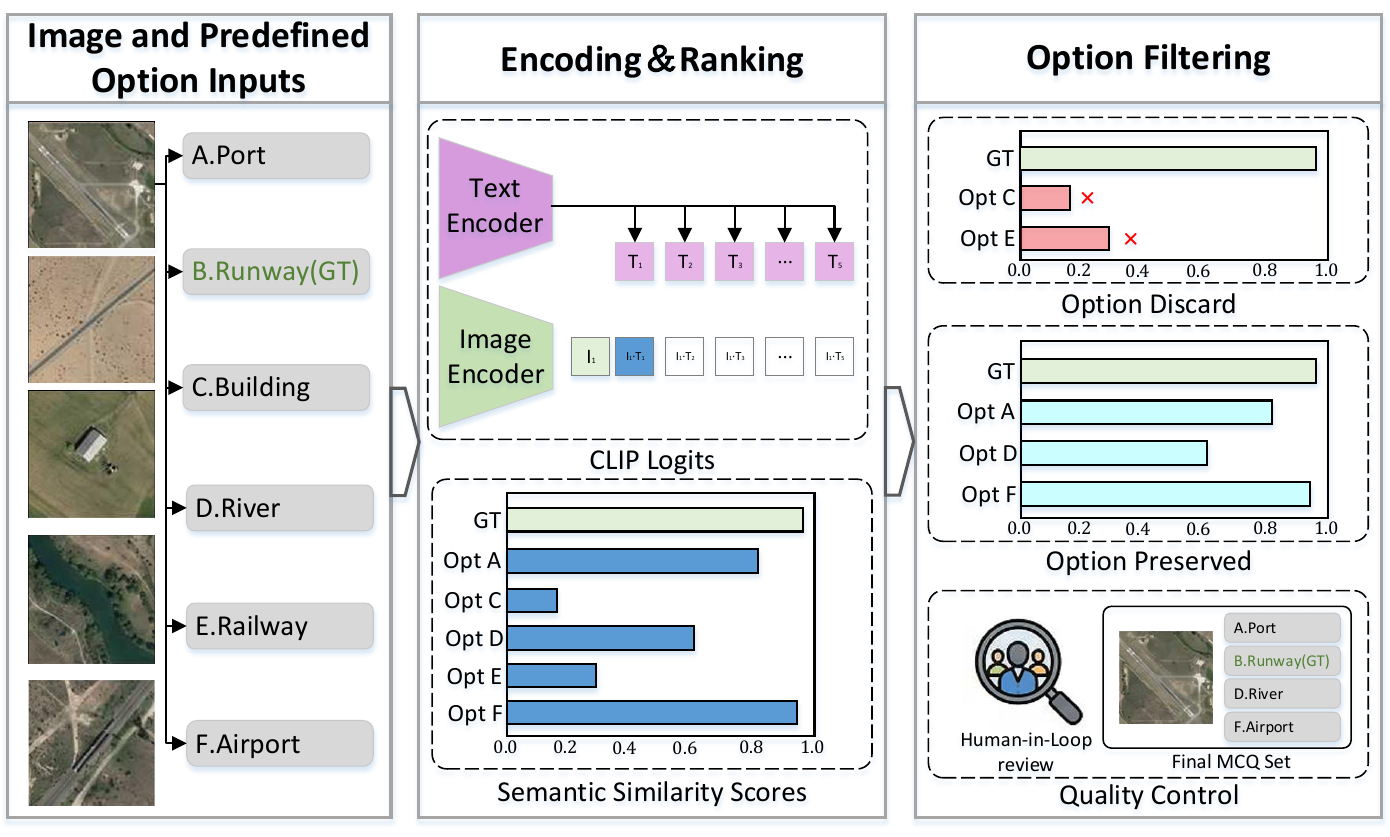}
\caption{Illustration of the SDFS. SDFS ranks candidate options according to image-text similarity and filters out overly easy or overly ambiguous distractors, thereby selecting plausible alternatives with moderate difficulty for MCQ construction.}
\label{fig4}
\end{figure}

In constructing MCQs, distractor quality directly affects both evaluation effectiveness and task difficulty~\cite{ref-48}. Effective distractors should be moderately difficult: they should be plausible to avoid trivial elimination, yet sufficiently distinct from the correct answer to preserve clear discriminability. Therefore, we adopt SDFS to filter the option pool with a reasonable difficulty range, as illustrated in Fig.~\ref{fig4}.

3.1) Semantic similarity computation. For each image, we employ a pretrained CLIP-based~\cite{ref-11} vision-language encoder to extract the visual representation of the image and the textual representations of a predefined option pool. Cosine similarity is then computed between the image feature and each option to establish the image-option similarity ranking. Importantly, these CLIP-based similarity scores are used only for candidate ranking and filtering, rather than for final label prediction. The correct answer for each MCQ is always inherited from the original ground-truth (GT) annotation.

3.2) Difficulty-aware distractor filtering. Based on the ranked candidates, SDFS further refines distractor selection by controlling the difficulty of each multiple-choice item. In this stage, candidates at both extremes of the ranking are discarded, for low-similarity options are too easy to eliminate, whereas high-similarity options are also controlled because they may be overly confusable with the GT option. Therefore, distractors are selected from the middle portion of the predefined option pool. The GT option is always preserved. If the GT appears at either extreme, it is retained, and the nearest non-GT candidate on that side is removed instead. This strategy yields distractors with moderate yet well-balanced difficulty.

3.3) Human expert verification. To further improve quality, we introduce a Human-in-the-Loop review stage after option filtering. Domain experts examine the preserved options and revise those that remain semantically inappropriate, visually implausible, or excessively ambiguous. This quality-control step ensures that the final distractors are appropriately challenging with respect to the GT, thereby improving the reliability and consistency of the final MCQ set.

Through this similarity-based distractor filtering pipeline, RRS-10K alleviates the common problem of overly simple distractors in conventional datasets while preserving the reliability of the original GT. Since CLIP is used solely for similarity-based option filtering rather than label prediction, the benchmark can more faithfully evaluate the fine-grained recognition, contextual understanding, and reasoning capabilities of RS-VLMs under rare-scene conditions.
\subsection{Hierarchical Task Taxonomy}

\begin{figure*}[!t]
\includegraphics[width=\textwidth]{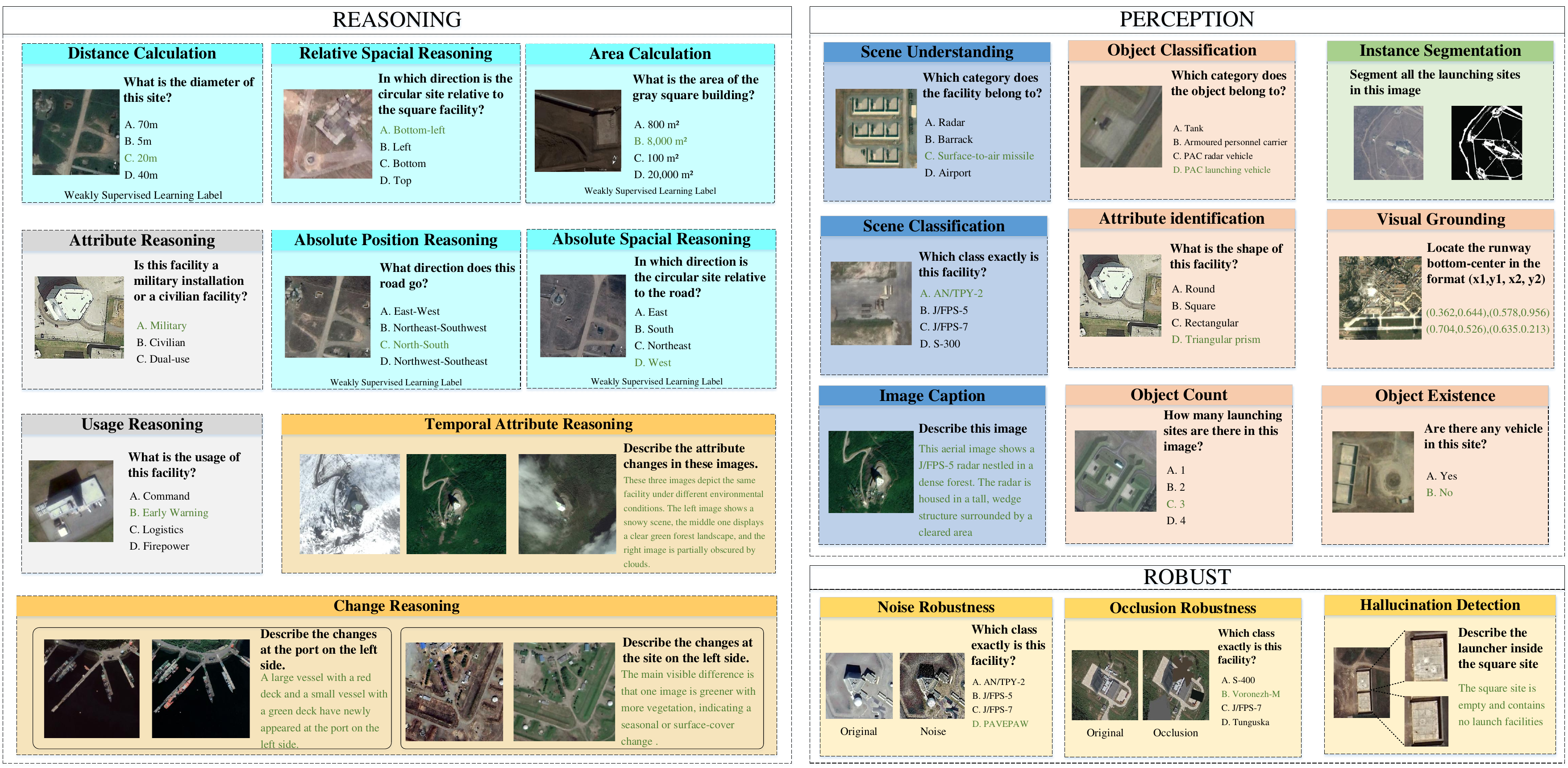}
\caption{Qualitative examples of fine-grained tasks in RRS-10K, organized by Reasoning, Perception, and Robustness dimensions. Green text indicates GT answers.\label{fig5}}
\end{figure*}

Hierarchical task taxonomy organizes benchmark tasks according to model capabilities, enabling comprehensive evaluation of VLM capabilities~\cite {ref-34,ref-39}. As shown in Fig.~\ref{fig5}, we constructed a hierarchical task taxonomy covering three capability dimensions, six sub-dimensions, and 20 leaf tasks for rare remote sensing scenes.

\unskip

\subsubsection{Perception}

\begin{figure}[H]
\includegraphics[width=\columnwidth]{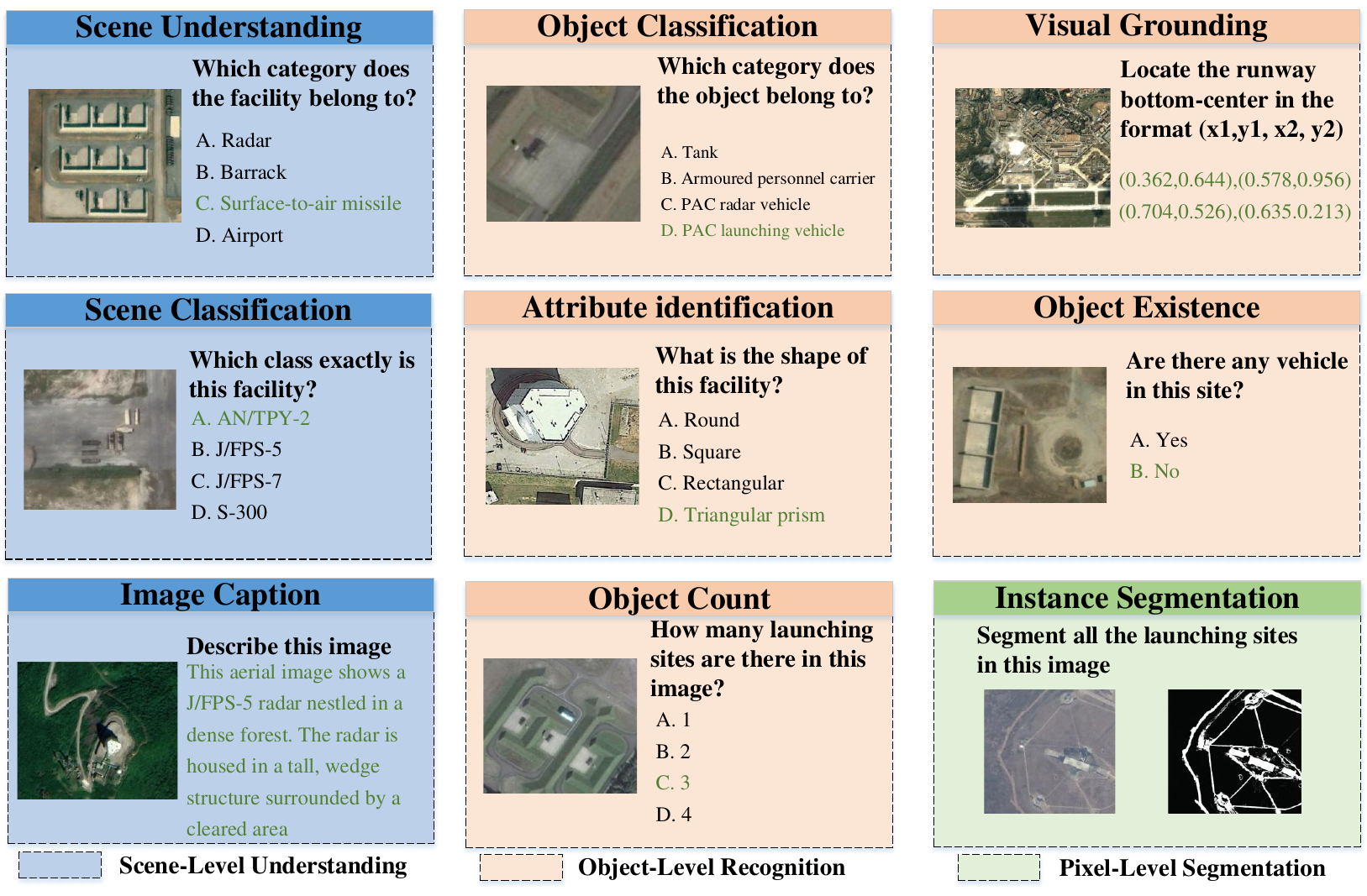}
\caption{Perception Task Visualization. The text in green indicates the GT answer.}
\label{fig6}
\end{figure}

The perception dimension assesses fundamental visual understanding in rare remote sensing imagery. Given the multi-scale characteristic of remote sensing scenes, we further organize perception into three sub-dimensions, namely image-level understanding, object-level recognition, and pixel-level segmentation, as illustrated in Fig.~\ref{fig6}.

Image-level Understanding evaluates the model’s global interpretation of rare remote sensing scenes. This sub-dimension targets global semantics, functional context, and overall spatial organization of a scene. Representative leaf tasks include scene understanding, scene classification, and image captioning. We collected open-source military scene images covering 6 major categories and 57 subcategories as the test scenarios. It includes (1) Scene Understanding, a coarse-grained thematic recognition task and requires the model to identify the primary scene type and dominant semantic theme of the image; the evaluated scenes cover 6 categories of open-source military imagery, namely launch sites, training grounds, radar facilities, ports, barracks, and airports; (2) Scene Classification, which further assesses discrimination among scene subtypes and functionally similar categories; taking radar scenes as an example, this category includes subtypes such as PAVEPAW radar. (3) Image Captioning, which requires the model to generate natural-language descriptions of salient objects and spatial relationships in the image.

Object-level Recognition shifts the evaluation focus from global scene understanding to instance-centered perception. This sub-dimension assesses the model’s capacity for identifying, localizing, counting, and describing rare targets embedded in complex backgrounds. Specifically, it includes: (1) Object Counting, which assesses target enumeration under long-tail distributions and scale variation, particularly for target arrangements such as multiple launch units distributed in parallel rows; (2) Visual Grounding, which requires the model to localize a referred entity according to a textual description and output its position in normalized bounding-box coordinates, where the reference may involve directional, structural, and contextual cues, such as the rightmost projecting pier associated with a moored warship; (3) Attribute Identification, which evaluates the model's ability to recognize discriminative object-level properties and structural characteristics, such as color, shape, and appearance; (4) Object Existence, which tests whether the model can reliably determine the presence or absence of a specified target or target configuration in the image, for example, whether three long projecting docks are visible on a designated side of the port.

For the construction of the Visual Grounding benchmark, concise referring expressions are first generated by GPT-5.4~\cite{ref-45}, after which Grounding DINO~\cite{ref-46} is employed to produce candidate bounding boxes, then verified and manually corrected by human annotators. In this way, the task evaluates the model’s ability to align referring expressions with target regions under complex remote-sensing scenes.

Pixel-level Segmentation extends from object-level perception to boundary-aware understanding. This sub-dimension evaluates pixel-level delineation of rare targets. During benchmark construction, concise target descriptions are first generated by GPT-5.4~\cite{ref-45} from seed annotations, then used as prompts for SAM-3~\cite{ref-47} to obtain candidate masks, and finally verified and corrected by human annotators. In this way, the task evaluates the model’s capacity to align referring expressions with target regions under complex scenes.

\subsubsection{Reasoning}

\begin{figure}[!t]
\includegraphics[width=\columnwidth]{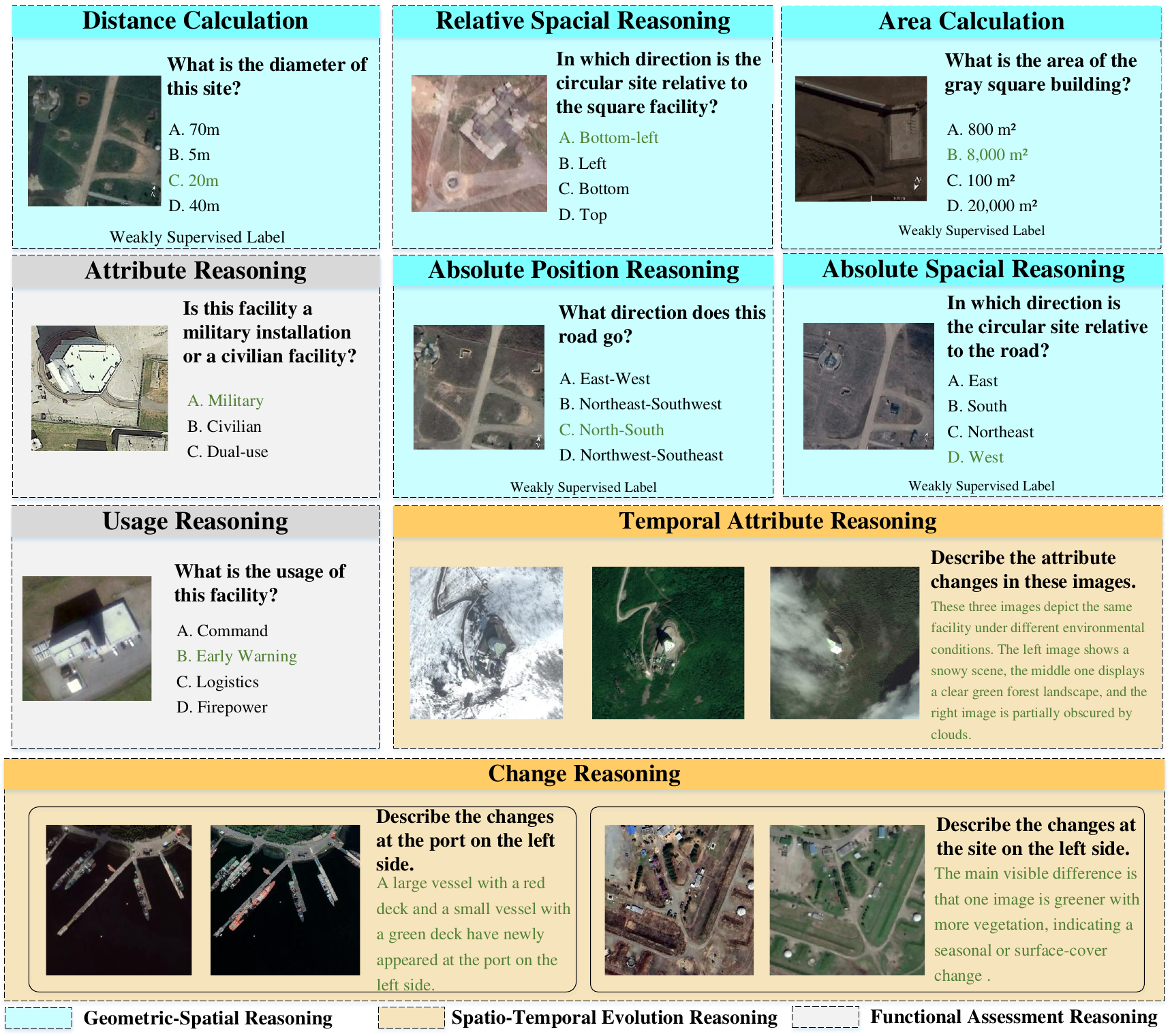}
\caption{Reasoning Task Visualization. We incorporate weak supervision signals inherent in raw images to evaluate models' high-order spatial reasoning capabilities. The text in green indicates the correct answer.\label{fig7}}
\end{figure}

In rare remote sensing image interpretation scenes, targets are often small, sparsely distributed, and embedded in large-scale backgrounds. Reliable interpretation therefore, relies on the ability to infer geometric relations, functional semantics, and temporal evolution from limited visual cues. As illustrated in Fig.~\ref{fig7}, we divide reasoning into three sub-dimensions: geometric-spatial reasoning, functional assessment reasoning, and spatio-temporal evolution reasoning. To support spatially grounded inference, RRS-10K preserves weak-supervision cues inherent in original imagery, including directional and metric-scale information.

Geometric-Spatial Reasoning targets the inference of directional structure, relative configuration, and measurable spatial extent from rare remote sensing scenes. This sub-dimension focuses on spatially grounded reasoning over target layout. It includes (1) Absolute Position Reasoning, which requires the model to infer the absolute geographic orientation of structures, such as roads, bridges, or runways; (2) Absolute Spatial Reasoning, which determines the absolute directional relation of one target with respect to another under geo-referenced scene cues; (3) Relative Spatial Reasoning, which assesses local positional relations between targets, such as left, right, above, below, or upper-left; (4) Distance Calculation, which evaluates the model’s ability to estimate physical length, interval, or diameter from image content and scale cues; (5) Area Calculation, which assesses estimation of the physical area or spatial extent of a target region or facility.

Functional Assessment Reasoning requires the model to infer functional semantics from visual appearance and scene context. In rare remote sensing images, different facilities may exhibit similar geometric layouts while serving different purposes, including (1) Usage Reasoning, which requires the model to infer the principal operational role of a facility, such as logistics support, firepower strike, reconnaissance, or training; (2) Attribute Reasoning, which assesses semantic scene and facility attributes, such as civilian, military, or dual-use, from structural configurations and contextual environments.

Spatio-temporal Evolution Reasoning targets temporal inference beyond single-image interpretation. This sub-dimension characterizes the temporal understanding of environmental states and facility evolution across time. It includes (1) Temporal Attribute Reasoning, which requires the model to infer temporal or environmental attributes reflected in imagery, such as seasonal conditions, vegetation state, or other acquisition-time cues; (2) Change Reasoning, which assesses temporal change identification and description across images, including the appearance, relocation, or structural modification.

\subsubsection{Robustness}

\begin{figure}[!t]
\centering
\includegraphics[width=\columnwidth]{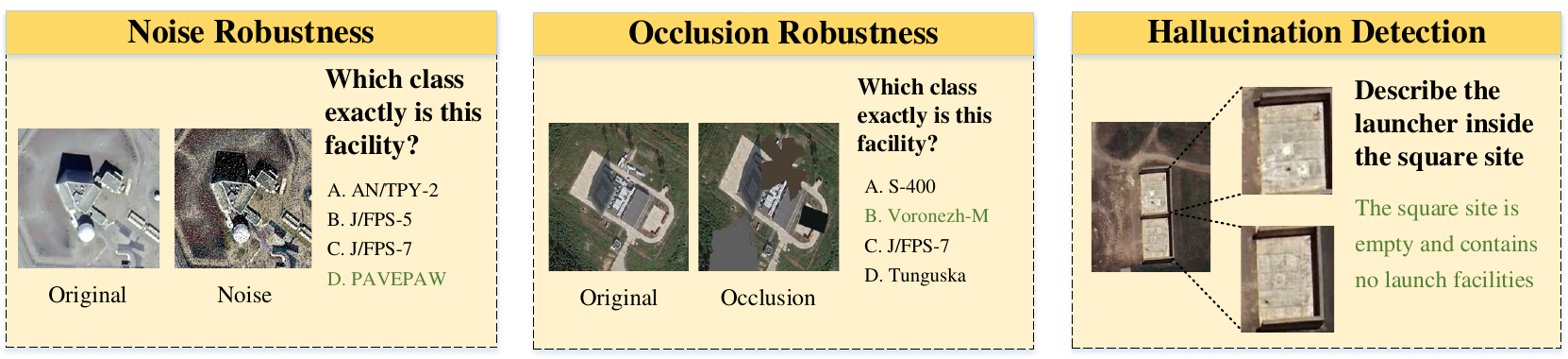}
\caption{Robustness Task Visualization. The text in green indicates the GT answer.\label{fig8}}
\end{figure}

The robustness dimension assesses the stability and reliability under imperfect observation conditions. As a result, model predictions are particularly vulnerable to image degradation, partial target visibility, and false-positive judgments. To assess this capability systematically, we divide robustness into 3 sub-dimensions, as illustrated in Fig.~\ref{fig8}: noise robustness, occlusion robustness, and hallucination detection.

Noise Robustness evaluates models' performance under image degradation. In rare remote sensing scenes, mild perturbations may distort local appearance cues. To simulate realistic observation disturbances, we introduce common image corruptions, including Gaussian noise and salt-and-pepper noise, to assess whether the model can still preserve reliable judgments.

Occlusion Robustness measures prediction reliability when critical regions are partially missing. If the model relies excessively on fragile local cues, partial masking may lead to degradation. To evaluate this capability, we apply random occlusion to local image regions and examine whether the model can still identify targets.

Hallucination Detection is designed to assess models' ability to suppress false-positive predictions from misleading prompts. To construct this setting, we first provide manually annotated data specifying facilities that are truly absent from a scene, and GPT-5.4~\cite{ref-45} is used to generate hallucination-oriented questions. Negative questions are mixed with normal positive questions to form deceptive evaluation samples. In this way, the task examines the model's capacity to distinguish true target presence from plausible but unsupported visual cues.


\section{Evaluation}
\label{sec:evaluation}

\subsection{Experimental Setup}

To comprehensively evaluate the capabilities of VLMs, we conducted extensive experiments on the RRS-10K benchmark using 52 representative models, including 43 VLMs and 9 referring segmentation models. The evaluated models span a wide range of parameter sizes, from 7B to 235B, and include both open-source and proprietary models. 

Experiments for models with 8B parameters or fewer were conducted on a Linux Ubuntu 22.04 platform equipped with an NVIDIA vGPU instance configured with 48 GB of GPU memory. Larger-scale and proprietary models were evaluated using inference via an asynchronous API.

\subsection{Evaluation Metrics}

Given the various task forms in RRS-10K, we adopt task-specific evaluation metrics to assess model performance across classification, generation, localization, and segmentation tasks.

(1) MCQs. For MCQ tasks, we use Accuracy as the primary evaluation metric:
\begin{equation}
\mathrm{Acc} = \frac{1}{N}\sum_{i=1}^{N}\mathbb{I}(\hat{y}_i = y_i),
\end{equation}
where $N$ denotes the total number of test samples, $\hat{y}_i$ denotes the predicted answer for the $i$-th sample, $y_i$ denotes the corresponding GT answer, and $\mathbb{I}(\cdot)$ is the indicator function.

(2) Open-ended tasks. For image captioning and other generative tasks, we adopt BLEU-1/2/3/4, ROUGE-L, and BERTScore~\cite{ref-49},~\cite{ref-50},~\cite{ref-51} to evaluate lexical overlap, sequence-level similarity, and semantic consistency between generated outputs and reference texts.

BLEU-$n$ is computed as
\begin{equation}
\mathrm{BLEU}\text{-}n = \mathrm{BP} \cdot \exp\left(\frac{1}{n}\sum_{k=1}^{n}\log p_k\right),
\end{equation}
where $\mathrm{BP}$ denotes the brevity penalty and $p_k$ denotes the modified precision of the $k$-gram.

ROUGE-L is based on the longest common subsequence (LCS) between the generated text $\hat{Y}$ and the reference text $Y$:
\begin{equation}
P_{\mathrm{LCS}} = \frac{\mathrm{LCS}(\hat{Y},Y)}{|\hat{Y}|}, \qquad
R_{\mathrm{LCS}} = \frac{\mathrm{LCS}(\hat{Y},Y)}{|Y|},
\end{equation}
\begin{equation}
\mathrm{ROUGE}\text{-}L = \frac{2P_{\mathrm{LCS}}R_{\mathrm{LCS}}}{P_{\mathrm{LCS}} + R_{\mathrm{LCS}}},
\end{equation}

To further evaluate semantic similarity beyond lexical matching, we use BERTScore:
\begin{equation}
P_{\mathrm{BERT}} = \frac{1}{|\hat{Y}|}\sum_{\hat{y}\in \hat{Y}}\max_{y\in Y}\cos(\mathbf{e}_{\hat{y}}, \mathbf{e}_{y}),
\end{equation}
\begin{equation}
R_{\mathrm{BERT}} = \frac{1}{|Y|}\sum_{y\in Y}\max_{\hat{y}\in \hat{Y}}\cos(\mathbf{e}_{y}, \mathbf{e}_{\hat{y}}),
\end{equation}
\begin{equation}
\mathrm{BERTScore}\text{-}F_1 = \frac{2P_{\mathrm{BERT}}R_{\mathrm{BERT}}}{P_{\mathrm{BERT}} + R_{\mathrm{BERT}}},
\end{equation}
where $\mathbf{e}_{\hat{y}}$ and $\mathbf{e}_{y}$ denote the contextual embeddings of tokens in the generated and reference texts, respectively.

(3) Visual grounding tasks.
For visual grounding tasks, we first compute the Intersection over Union (IoU) between the predicted bounding box $B_p$ and the GT bounding box $B_g$~\cite{ref-52}:
\begin{equation}
\mathrm{IoU} = \frac{|B_p \cap B_g|}{|B_p \cup B_g|},
\end{equation}

Following standard grounding protocols~\cite{ref-53}, grounding performance is then reported using threshold-based accuracy:
\begin{equation}
\mathrm{Acc}@\tau = \frac{1}{N}\sum_{i=1}^{N}\mathbb{I}(\mathrm{IoU}_i > \tau),
\end{equation}
where $\tau$ denotes the IoU threshold.

(4) Segmentation tasks.
For pixel-level segmentation tasks, we adopt mean Intersection over Union (mIoU) and Dice coefficient~\cite{ref-54,ref-55} to evaluate mask overlap quality:
\begin{equation}
\mathrm{mIoU} = \frac{1}{C}\sum_{c=1}^{C}\frac{TP_c}{TP_c + FP_c + FN_c},
\end{equation}
\begin{equation}
\mathrm{Dice} = \frac{2|M_p \cap M_g|}{|M_p| + |M_g|},
\end{equation}
where $C$ denotes the number of classes; $TP_c$, $FP_c$, and $FN_c$ denote the numbers of true-positive, false-positive, and false-negative pixels for class $c$, respectively; and $M_p$ and $M_g$ denote the predicted and GT masks.

\subsection{Main Results}

\subsubsection{Overall Results}

We first evaluate 43 representative open-source and proprietary VLMs on RRS-10K. Overall, the results show that high-performing open-source models have become broadly competitive with proprietary systems in rare remote sensing scene interpretation. Gemini-3-Flash achieves the highest average score, while Qwen3-VL-235B-Instruct ranks second and slightly surpasses GPT-4o. In addition, Qwen3-VL-8B and Qwen2.5-VL-72B-Instruct also remain highly competitive, suggesting that the performance gap between proprietary and open-source models has narrowed substantially under rare-scene zero-shot evaluation.

Model scaling is beneficial but not uniformly decisive. For example, Qwen3-VL-4B already outperforms many open-source models with parameter sizes ranging from 7B to 32B, while Qwen3-VL-8B-Instruct further surpasses several substantially larger counterparts. This trend suggests that, besides model size, data quality, multimodal alignment, and instruction tuning also play critical roles in rare-scene interpretation.

Task-level results are reported in Table~\ref{tab:model_comparison_colored}, providing a comparison across 20 leaf tasks.

\makeatletter
\setlength{\@fptop}{0pt}
\setlength{\@fpbot}{0pt plus 1fil}
\setlength{\@dblfptop}{0pt}
\setlength{\@dblfpbot}{0pt plus 1fil}
\makeatother
\begin{table*}[!t]
\centering
\scriptsize
\renewcommand{\arraystretch}{1.00}
\setlength{\tabcolsep}{3.6pt}

\definecolor{bestcolor}{RGB}{248,215,218}
\definecolor{secondcolor}{RGB}{219,233,251}
\definecolor{avgcolor}{RGB}{226,239,218}
\definecolor{groupgray}{RGB}{235,235,235}

\renewcommand{\best}[1]{\cellcolor{bestcolor}\textbf{\textcolor{black}{#1}}}
\renewcommand{\second}[1]{\cellcolor{secondcolor}\textbf{\textcolor{black}{#1}}}

\renewcommand{\blockheader}[1]{%
\rowcolor{groupgray}
\multicolumn{21}{@{}l}{\textbf{#1}}\\[-0.2em]
}

\renewcommand{\seriesheader}[1]{%
\multicolumn{21}{@{}l}{\textbf{#1}}\\[-0.15em]
}

\renewcommand{\groupsep}{%
\noalign{\vskip 0.45em}
\cdashline{1-21}[1.2pt/2.2pt]
\noalign{\vskip 0.45em}
}

\caption{Detailed task-level performance comparison of 43 representative VLMs on RRS-10K across 20 leaf tasks.}
\label{tab:model_comparison_colored}
\resizebox{\textwidth}{!}{%
\begin{tabular}{@{}lrrrrrrrrrrrrrrrrrrrr@{}}
\toprule
\textbf{Model} & \textbf{Avg} & \textbf{SU} & \textbf{SC} & \textbf{ID} & \textbf{OC} & \textbf{VG} & \textbf{AR} & \textbf{OE} & \textbf{API} & \textbf{RSI} & \textbf{ASI} & \textbf{AC} & \textbf{DC} & \textbf{UI} & \textbf{AI} & \textbf{TAI} & \textbf{CI} & \textbf{NR} & \textbf{OR} & \textbf{HD} \\
\midrule

\blockheader{Proprietary Models}
GPT-5.4 & 64.04 & 60.98 & 41.15 & 59.61 & 59.28 & \best{49.89} & 64.18 & 85.71 & \best{50.32} & \best{93.39} & \second{74.84} & 64.04 & 80.60 & 39.66 & 30.28 & 55.44 & \best{94.43} & 53.88 & 61.27 & 88.65 \\
GPT-4o & 69.60 & \second{80.81} & 77.19 & 90.46 & 61.41 & 14.93 & 63.54 & 66.31 & 32.41 & 83.58 & 67.16 & 69.60 & 80.17 & \second{63.75} & 73.13 & 50.32 & 83.08 & \best{79.96} & \second{81.76} & 94.43 \\
Claude Sonnet 4.6 & 62.58 & 58.00 & 69.51 & 85.95 & 36.89 & \second{25.80} & 59.06 & 74.84 & 38.59 & 88.49 & 70.79 & 62.58 & 72.71 & 40.94 & 49.89 & 55.44 & 86.30 & 67.78 & 70.06 & 87.37 \\
Gemini-3-Flash & \best{71.43} & \best{81.66} & \best{87.85} & 84.84 & 59.28 & 0.00 & \best{71.22} & \second{85.93} & \second{48.83} & 89.13 & 74.41 & \second{71.43} & 75.91 & \best{66.74} & 87.42 & 51.39 & 87.37 & 72.41 & \best{83.15} & 86.94 \\
Grok-4.1 & 54.27 & 48.61 & 65.88 & 87.81 & 52.45 & 12.58 & 51.17 & 46.06 & 35.61 & 60.77 & 53.09 & 54.27 & 66.52 & 34.97 & 41.36 & 38.38 & 73.66 & 52.24 & 54.46 & 91.86 \\
\groupsep
\blockheader{Open-source Models}
Qwen3-VL-2B & 44.15 & 41.58 & 49.68 & 90.35 & 25.37 & 0.00 & 40.30 & 55.01 & 14.50 & 66.74 & 27.29 & 10.45 & 34.97 & 10.02 & 78.04 & 25.80 & 56.53 & 49.68 & 53.97 & 95.72 \\
Qwen3-VL-4B & 62.82 & 49.25 & 55.65 & \second{90.72} & 46.70 & 0.21 & 55.44 & 52.24 & 40.30 & 81.66 & 63.54 & 64.61 & 72.71 & 26.44 & 89.34 & 38.59 & \second{91.22} & 66.06 & 67.70 & \second{98.72} \\
Qwen2.5-VL-7B-Instruct & 63.90 & 68.95 & 64.39 & 90.07 & 51.29 & 0.00 & 56.20 & 53.42 & 30.54 & 78.80 & 58.33 & 63.11 & 76.28 & 34.62 & 83.55 & 29.55 & 85.29 & 67.78 & 73.93 & 97.84 \\
Qwen3-VL-8B & 67.64 & 61.83 & 68.02 & 90.57 & 62.47 & 0.00 & 59.28 & 60.98 & 39.45 & 82.52 & 58.21 & 66.95 & 78.04 & 33.26 & 91.90 & 35.39 & 89.29 & \second{73.60} & 77.36 & 96.15 \\
Qwen3-VL-8B-Instruct & 66.31 & 62.96 & 72.71 & 90.59 & 59.87 & 0.00 & 59.40 & 51.28 & 37.42 & 82.01 & 57.91 & 64.61 & 75.64 & 34.12 & 91.88 & 35.55 & 87.10 & 70.26 & 74.79 & 97.62 \\
Qwen3-VL-8B-Thinking & 60.82 & 45.61 & 61.19 & 90.20 & 55.58 & 0.22 & 57.48 & 47.44 & 43.66 & 84.33 & 62.50 & 57.48 & 76.71 & 35.04 & 69.87 & 44.33 & 66.45 & 61.31 & 63.63 & 98.49 \\
Qwen3-VL-30B-A3B-Instruct & 65.65 & 54.82 & 78.25 & 90.24 & 52.79 & 0.00 & 58.97 & 48.48 & 27.53 & 88.22 & 72.27 & 64.25 & 77.26 & 41.11 & 84.25 & 54.00 & 88.73 & 65.44 & 65.99 & 97.28 \\
Qwen3-VL-30B-A3B-Thinking & 64.20 & 64.24 & 67.80 & 88.82 & 55.15 & 0.00 & 64.10 & 45.73 & 38.71 & 86.51 & 64.96 & 66.10 & 73.50 & 33.97 & 78.21 & 44.75 & 81.72 & 61.96 & 66.20 & 98.27 \\
Qwen3-Omni-30B-A3B-Captioner & 63.66 & 56.53 & \second{78.68} & 89.53 & 42.70 & 0.43 & 51.45 & 51.66 & 24.83 & 81.58 & 65.47 & 66.96 & 75.72 & 37.53 & 89.02 & 47.01 & 90.03 & 63.66 & 65.00 & 97.12 \\
Qwen3-Omni-30B-A3B-Instruct & 64.83 & 69.16 & 68.23 & 90.30 & 46.35 & 0.22 & 47.79 & 55.34 & 24.01 & 83.60 & 66.30 & 65.36 & 78.07 & 37.34 & 92.14 & \second{56.12} & 90.58 & 63.73 & 66.21 & 95.48 \\
Qwen3-Omni-30B-A3B-Thinking & 67.29 & 68.52 & 72.28 & 90.26 & 60.09 & 3.70 & 61.39 & 47.01 & 38.71 & 86.12 & 66.45 & 66.95 & 80.77 & 43.16 & 74.79 & 49.89 & 82.33 & 69.54 & 72.82 & 97.58 \\
Qwen2.5-VL-32B-Instruct & 63.39 & 70.45 & 59.06 & 87.28 & 55.15 & 0.64 & 56.41 & 55.56 & 26.02 & 85.44 & 64.32 & 67.38 & 77.48 & 44.23 & 70.52 & 38.33 & 88.29 & 61.10 & 65.74 & 96.44 \\
Qwen3-VL-32B-Instruct & 67.16 & 80.51 & 60.55 & 90.36 & 60.73 & 0.00 & \second{68.16} & 62.18 & 40.00 & 85.65 & 65.17 & 58.64 & 80.98 & 38.25 & 81.84 & 49.25 & 87.74 & 61.64 & 67.60 & 98.06 \\
Qwen3-VL-32B-Thinking & 63.56 & 65.52 & 62.26 & 90.09 & 58.80 & 0.00 & 61.11 & 51.50 & 48.39 & 87.58 & 67.09 & 53.73 & 73.72 & 32.69 & 73.93 & 35.19 & 74.30 & 63.15 & 68.24 & \best{99.35} \\
Qwen2-VL-72B-Instruct & 62.66 & 49.04 & 65.97 & 90.06 & 51.29 & 12.24 & 60.09 & 64.06 & 45.14 & 86.83 & 65.18 & 69.42 & 80.89 & 41.30 & 55.32 & 36.06 & 81.17 & 62.72 & 65.93 & 93.65 \\
Qwen2.5-VL-72B-Instruct & 67.32 & 79.01 & 61.52 & 90.50 & \second{62.50} & 7.29 & 61.33 & 58.87 & 37.70 & 89.80 & 73.37 & \best{72.82} & 80.47 & 38.12 & 55.26 & 45.26 & 89.29 & 66.71 & 75.55 & 97.33 \\
Qwen3-VL-235B-Instruct & \second{69.63} & 70.53 & 64.49 & \best{90.89} & 56.31 & 0.00 & 64.22 & 56.30 & 37.71 & \second{90.95} & \best{77.02} & 70.00 & \best{85.41} & 48.51 & 75.83 & \best{62.60} & 88.99 & 71.69 & 70.36 & 96.15 \\
Qwen3-VL-235B-Thinking & 66.08 & 68.93 & 67.82 & 90.13 & \best{63.52} & 0.86 & 64.32 & 48.50 & 47.31 & 88.44 & 64.53 & 68.02 & \second{85.34} & 30.29 & 62.84 & 46.47 & 74.19 & 67.86 & 68.70 & 98.12 \\
\groupsep
GLM-4.5V & 62.94 & 59.90 & 61.86 & 90.41 & 47.60 & 0.66 & 61.38 & 65.10 & 38.12 & 89.33 & 71.53 & 54.32 & 75.13 & 31.93 & 80.77 & 43.48 & 75.61 & 59.87 & 63.23 & 97.01 \\
GLM-4.6V & 64.78 & 66.81 & 56.93 & 89.50 & 43.13 & 0.21 & 61.32 & 66.24 & 44.30 & 88.65 & 71.58 & 50.53 & 76.28 & 34.83 & 81.62 & 48.39 & 87.59 & 62.93 & 65.45 & 98.06 \\
\groupsep
DeepSeek-VL2-tiny & 51.46 & 68.87 & 75.48 & 89.92 & 39.23 & 1.28 & 50.11 & 28.78 & 27.29 & 61.62 & 49.68 & 3.20 & 47.55 & 19.40 & 65.25 & 31.13 & 67.02 & 64.87 & 63.52 & 86.94 \\
DeepSeek-vl-7B & 51.64 & 45.20 & 33.69 & 88.54 & 39.02 & 6.18 & 48.40 & 56.29 & 28.14 & 58.00 & 49.04 & 26.23 & 59.28 & 35.82 & 71.43 & 44.78 & 58.46 & 53.88 & 61.91 & 96.15 \\
DeepSeek-VL2 & 36.82 & 25.72 & 49.46 & 90.18 & 19.79 & 10.56 & 59.51 & 22.07 & 26.60 & 76.34 & 57.61 & 24.21 & 46.62 & 6.05 & 11.68 & 39.73 & 62.68 & 31.39 & 30.25 & 65.96 \\
DeepSeek-OCR & 4.61 & 0.21 & 5.97 & 89.06 & 0.64 & 0.00 & 4.27 & 5.34 & 2.15 & 0.43 & 0.00 & 0.43 & 3.42 & 2.14 & 1.07 & 0.00 & 0.00 & 10.45 & 13.95 & 12.74 \\
\groupsep
PaddleOCR-VL & 9.56 & 1.28 & 20.68 & 81.48 & 6.22 & 0.00 & 7.48 & 7.26 & 2.37 & 31.91 & 10.47 & 0.00 & 0.43 & 2.99 & 0.21 & 7.49 & 0.00 & 12.39 & 11.37 & 26.13 \\
PaddleOCR-VL-1.5 & 3.73 & 0.86 & 12.15 & 75.68 & 3.90 & 0.00 & 3.42 & 2.74 & 0.00 & 4.71 & 0.22 & 0.00 & 0.21 & 0.43 & 0.23 & 2.88 & 0.00 & 9.59 & 6.87 & 2.16 \\
\groupsep
LLaVA-1.5-7b & 44.18 & 37.10 & 27.93 & 88.87 & 32.41 & 7.25 & 40.72 & 82.30 & 28.57 & 62.47 & 49.47 & 2.13 & 24.09 & 23.67 & 88.70 & 28.57 & 56.10 & 45.58 & 43.99 & 76.23 \\
LLaVA-NeXT-Video-7B & 14.34 & 0.00 & 16.84 & 89.29 & 10.66 & 0.00 & 0.21 & 14.07 & 11.51 & 6.40 & 20.68 & 0.00 & 0.43 & 0.00 & 0.00 & 13.43 & 15.63 & 31.47 & 27.58 & 45.18 \\
LLaVA-v1.6-mistral-7b & 36.25 & 33.69 & 41.15 & 89.70 & 19.83 & 21.32 & 23.24 & \best{86.78} & 18.12 & 52.67 & 47.33 & 0.00 & 9.59 & 12.15 & 4.48 & 40.09 & 50.75 & 43.21 & 38.41 & 86.08 \\
LLaVA-OneVision-1.5-8B & 21.53 & 1.71 & 40.94 & 89.25 & 10.66 & 0.00 & 15.35 & 14.93 & 18.12 & 51.17 & 55.44 & 0.00 & 1.92 & 4.05 & 0.43 & 12.79 & 18.63 & 31.47 & 27.58 & 45.18 \\
VideoLLaMA2-7B & 52.47 & 67.38 & 51.17 & 88.85 & 22.17 & 5.54 & 47.97 & 82.30 & 31.13 & 67.59 & 31.98 & 30.70 & 53.30 & 29.85 & 81.02 & 37.74 & 63.38 & 57.00 & 57.08 & 71.31 \\
\groupsep
MiniCPM-V-2-6 & 55.18 & 69.08 & 42.00 & 89.38 & 43.50 & 2.13 & 55.86 & 52.67 & 31.34 & 51.39 & 21.54 & 42.86 & 71.00 & 39.87 & \second{96.16} & 33.69 & 82.23 & 51.94 & 56.65 & 98.29 \\
MiniCPM-V-4-5 & 26.95 & 4.05 & 41.58 & 90.51 & 20.90 & 7.04 & 27.08 & 19.83 & 18.55 & 52.67 & 42.22 & 1.07 & 31.56 & 4.26 & 6.82 & 14.71 & 44.33 & 31.90 & 28.54 & 61.88 \\
\groupsep
Internlm-xcomposer2-VL-1.8B & 44.51 & 56.08 & 24.31 & 88.31 & 37.10 & 1.28 & 38.38 & 47.55 & 20.90 & 50.11 & 26.87 & 27.29 & 35.82 & 15.78 & 70.15 & 40.51 & 47.75 & 51.19 & 55.47 & 94.22 \\
Internlm-xcomposer2-vl-7B & 21.53 & 1.71 & 40.94 & 83.25 & 10.66 & 0.00 & 15.35 & 14.93 & 18.12 & 51.17 & 55.44 & 0.00 & 1.92 & 4.05 & 0.43 & 12.79 & 18.63 & 31.47 & 27.58 & 45.18 \\
\groupsep
Phi-3.5-Vision & 50.43 & 49.25 & 44.14 & 89.20 & 26.87 & 2.77 & 46.06 & 42.64 & 29.21 & 64.61 & 39.87 & 38.81 & 68.87 & 20.90 & 94.67 & 35.61 & 88.87 & 42.56 & 43.67 & 95.50 \\
Phi3-Vision & 52.37 & 33.05 & 55.01 & 84.96 & 41.58 & 0.43 & 49.68 & 52.67 & 25.80 & 68.66 & 34.75 & 48.61 & 71.86 & 26.65 & \best{96.80} & 29.85 & 71.73 & 47.63 & 48.18 & 96.79 \\
Phi-4-Multimodal & 60.22 & 71.43 & 58.21 & 89.77 & 46.27 & 5.33 & 56.29 & 74.84 & 24.31 & 75.48 & 52.67 & 58.64 & 65.25 & 32.41 & 95.10 & 34.97 & 79.66 & 54.31 & 60.62 & 88.87 \\
\groupsep
\rowcolor{avgcolor}
\textbf{Model Average} & 52.06 & 50.02 & 53.97 & 88.04 & 40.89 & 4.67 & 48.11 & 50.32 & 30.05 & 69.71 & 52.76 & 42.97 & 57.78 & 28.68 & 61.11 & 36.69 & 67.61 & 53.94 & 56.10 & 83.45 \\
\bottomrule
\end{tabular}%
}
\par\smallskip
\begin{minipage}{\textwidth}
\footnotesize
\raggedright
\textsuperscript{}The best and second-best results in each column are highlighted in
\begingroup\setlength{\fboxsep}{1pt}\colorbox{bestcolor}{\strut light red}\endgroup and\begingroup\setlength{\fboxsep}{1pt}\colorbox{secondcolor}{\strut light blue}\endgroup. SU, Scene Understanding; SC, Scene Classification; ID, Image Description; OC, Object Counting; VG, Visual Grounding; AR, Attribute Recognition; OE, Object Existence; API, Absolute Position Inference; RSI, Relative Spatial Inference; ASI, Absolute Spatial Inference; AC, Area Calculation; DC, Distance Calculation; UI, Usage Inference; AI, Attribute Inference; TAI, Temporal Attribute Inference; CI, Change Identification; NR, Noise Robustness; OR, Occlusion Robustness; HD, Hallucination Detection.
\end{minipage}
\end{table*}

The overall ranking is led by Gemini-3-Flash~\cite{ref-56}, but the advantage of proprietary models is no longer absolute at the top tier. Notably, Qwen3-VL-235B-Instruct~\cite{ref-57} achieves the second-highest average score and slightly surpasses GPT-4o~\cite{ref-58}, while Qwen3-VL-8B~\cite{ref-57} and Qwen2.5-VL-72B-Instruct~\cite{ref-59} also remain highly competitive. Although Claude Sonnet 4.6~\cite{ref-60} still performs solidly, the results overall suggest that leading open-source VLMs have already become broadly comparable to proprietary models for rare scene tasks.

Additionally, model scaling is beneficial but not uniformly decisive in rare zero-shot tasks. For instance, Qwen3-VL-4B already outperforms many open-source models with parameter sizes ranging from 7B to 32B, while Qwen3-VL-8B-Instruct further surpasses several substantially larger counterparts.

To provide a more compact and capability-oriented view, Table~\ref{tab:model_comparison} further reorganizes results into higher-level summaries. MCQ tasks are aggregated into Level-2 capability dimensions, while the open-ended results are reported separately using lexical overlap and semantic-similarity metrics.


\begin{table*}[!t]
\centering
\scriptsize
\renewcommand{\arraystretch}{1.00}
\setlength{\tabcolsep}{3.6pt}

\definecolor{bestcolor}{RGB}{248,215,218}
\definecolor{secondcolor}{RGB}{219,233,251}
\definecolor{avgcolor}{RGB}{226,239,218}
\definecolor{groupgray}{RGB}{235,235,235}

\renewcommand{\best}[1]{\cellcolor{bestcolor}\textbf{\textcolor{black}{#1}}}
\renewcommand{\second}[1]{\cellcolor{secondcolor}\textbf{\textcolor{black}{#1}}}

\renewcommand{\blockheader}[1]{%
\rowcolor{groupgray}
\multicolumn{18}{@{}l}{\textbf{#1}}\\[-0.2em]
}

\renewcommand{\seriesheader}[1]{%
\multicolumn{18}{@{}l}{\textbf{#1}}\\[-0.15em]
}

\renewcommand{\groupsep}{%
\noalign{\vskip 0.45em}
\cdashline{1-18}[1.2pt/2.2pt]
\noalign{\vskip 0.45em}
}

\caption{Capability-level performance comparison of 43 representative VLMs.}
\label{tab:model_comparison}
\resizebox{\textwidth}{!}{%
\begin{tabular}{@{}l*{17}{c}@{}}
\toprule
\textbf{FORMAT} & \multicolumn{7}{c}{\textbf{MCQ}} & \multicolumn{10}{c}{\textbf{Open-Ended}} \\
\cmidrule(lr){2-8} \cmidrule(lr){9-18}
\textbf{MODEL} & \textbf{Avg} & \textbf{ILU} & \textbf{OLR} & \textbf{GSI} & \textbf{FA} & \textbf{STE} & \textbf{RB} & \textbf{Bleu-1} & \textbf{Bleu-2} & \textbf{Bleu-3} & \textbf{Bleu-4} & \textbf{Rouge-1} & \textbf{Rouge-2} & \textbf{Rouge-L} & \textbf{BERT-P} & \textbf{BERT-R} & \textbf{BERT-F1} \\
\midrule

\blockheader{Proprietary Models}
GPT-5.4 & 64.04 & 56.44 & \best{64.77} & \best{72.64} & 34.97 & \second{74.93} & 67.93 & 31.90 & 12.42 & \best{5.48} & \best{2.64} & 35.69 & \best{13.90} & 20.10 & 59.09 & 60.14 & 59.61 \\
GPT-4o & 69.60 & \second{79.51} & 51.55 & 66.59 & \second{68.44} & 66.70 & \best{85.38} & \second{56.06} & \second{13.80} & \second{3.72} & \second{1.07} & 40.79 & 10.06 & 22.88 & \best{91.81} & 89.15 & 90.46 \\
Claude Sonnet 4.6 & 62.58 & 69.01 & 49.15 & 66.63 & 45.42 & 70.87 & 75.07 & 40.84 & 10.02 & 2.92 & 0.88 & 41.92 & 10.29 & 20.90 & 84.11 & 87.87 & 85.95 \\
Gemini-3-Flash & \best{71.43} & \best{81.45} & \second{54.10} & 71.94 & \best{77.08} & 69.38 & 80.84 & 51.05 & 10.45 & 1.26 & 0.39 & 5.65 & 0.99 & 4.99 & 89.29 & 80.84 & 84.84 \\
Grok-4.1 & 54.27 & 64.14 & 40.57 & 54.05 & 38.17 & 56.02 & 66.19 & 30.66 & 7.26 & 1.66 & 0.43 & 40.15 & 9.53 & 19.25 & 85.69 & 90.06 & 87.81 \\
\groupsep
\blockheader{Open-source Models}
Qwen3-VL-2B & 44.15 & 56.44 & 30.17 & 30.79 & 44.03 & 41.17 & 66.46 & 43.33 & 10.81 & 2.89 & 0.82 & 43.17 & 10.88 & 22.55 & 90.38 & 90.33 & 90.35 \\
Qwen3-VL-4B & 62.82 & 64.61 & 38.65 & 64.56 & 57.89 & 64.91 & 77.49 & 45.85 & 11.51 & 2.81 & 0.80 & \second{45.99} & 11.56 & \second{23.64} & 90.41 & \second{91.04} & \second{90.72} \\
Qwen2.5-VL-7B-Instruct & 63.90 & 71.83 & 40.23 & 61.41 & 59.09 & 57.42 & 79.85 & 45.46 & 9.24 & 2.07 & 0.46 & 38.55 & 7.88 & 20.47 & 90.83 & 89.32 & 90.07 \\
Qwen3-VL-8B & 67.64 & 72.02 & 45.68 & 65.03 & 62.58 & 62.34 & \second{82.37} & 47.39 & 11.81 & 3.08 & 0.93 & 44.95 & 11.25 & 23.62 & 90.43 & 90.74 & 90.57 \\
Qwen3-VL-8B-Instruct & 66.31 & 73.14 & 42.64 & 63.52 & 63.00 & 61.33 & 80.89 & 49.28 & 11.75 & 2.98 & 0.89 & 44.24 & 10.58 & 23.15 & 90.76 & 90.44 & 90.59 \\
Qwen3-VL-8B-Thinking & 60.82 & 64.46 & 40.18 & 64.94 & 52.46 & 55.39 & 74.48 & 45.21 & 10.32 & 2.40 & 0.66 & 43.86 & 10.02 & 22.65 & 89.64 & 90.78 & 90.20 \\
Qwen3-VL-30B-A3B-Instruct & 65.65 & 72.24 & 40.06 & 65.91 & 62.68 & 71.37 & 76.24 & 42.25 & 11.48 & 3.28 & 1.01 & 44.79 & 12.20 & 23.25 & 89.73 & 90.76 & 90.24 \\
Qwen3-VL-30B-A3B-Thinking & 64.20 & 71.27 & 41.25 & 65.96 & 56.09 & 63.24 & 75.48 & 40.67 & 9.60 & 2.41 & 0.66 & 43.62 & 10.33 & 21.92 & 87.30 & 90.41 & 88.82 \\
Qwen3-Omni-30B-A3B-Captioner & 63.66 & 72.10 & 36.56 & 62.91 & 63.28 & 68.52 & 75.26 & 35.92 & 9.33 & 2.33 & 0.63 & 41.58 & 10.82 & 20.99 & 88.68 & 90.40 & 89.53 \\
Qwen3-Omni-30B-A3B-Instruct & 64.83 & 73.13 & 37.43 & 63.47 & 64.74 & 73.35 & 75.14 & 41.42 & 10.95 & 3.01 & 0.85 & 43.82 & 11.62 & 22.82 & 89.87 & 90.73 & 90.30 \\
Qwen3-Omni-30B-A3B-Thinking & 67.29 & 74.59 & 43.05 & 67.80 & 58.98 & 66.11 & 79.98 & 43.97 & 10.55 & 2.56 & 0.72 & 44.72 & 10.72 & 22.81 & 89.59 & 90.95 & 90.26 \\
Qwen3-VL-32B-Thinking & 63.56 & 70.36 & 42.85 & 66.10 & 53.31 & 54.75 & 76.91 & 43.95 & 10.71 & 2.60 & 0.71 & 45.73 & 11.17 & 22.97 & 89.18 & 91.02 & 90.09 \\
Qwen2.5-VL-32B-Instruct & 63.39 & 70.05 & 41.94 & 64.13 & 57.38 & 63.31 & 74.43 & 35.91 & 8.54 & 2.33 & 0.58 & 39.91 & 9.51 & 19.98 & 85.41 & 89.25 & 87.28 \\
Qwen3-VL-32B-Instruct & 67.16 & 74.65 & 47.77 & 66.09 & 60.05 & 68.50 & 75.77 & 40.54 & 11.00 & 3.17 & 0.98 & 45.52 & \second{12.38} & 23.61 & 89.60 & 91.15 & 90.36 \\
Qwen2-VL-72B-Instruct & 62.66 & 66.93 & 46.92 & 69.49 & 48.31 & 58.62 & 74.10 & 47.03 & 10.04 & 2.44 & 0.64 & 40.09 & 8.70 & 20.89 & 90.68 & 89.47 & 90.06 \\
Qwen2.5-VL-72B-Instruct & 67.32 & 74.59 & 47.50 & 70.83 & 46.69 & 67.28 & 79.86 & 43.94 & 10.08 & 2.57 & 0.75 & 42.38 & 9.81 & 22.12 & 90.70 & 90.31 & 90.50 \\
Qwen3-VL-235B-Instruct & \second{69.63} & 73.89 & 44.21 & \second{72.22} & 62.17 & \best{75.80} & 79.40 & 45.07 & 11.84 & 3.34 & 1.06 & \best{46.06} & 12.15 & \best{24.39} & 90.48 & \best{91.31} & \best{90.89} \\
Qwen3-VL-235B-Thinking & 66.08 & 73.24 & 44.30 & 70.73 & 46.57 & 60.33 & 78.23 & 43.62 & 10.87 & 2.84 & 0.91 & 45.80 & 11.46 & 23.54 & 89.13 & \second{91.16} & 90.13 \\

\groupsep
GLM-4.5V & 62.94 & 68.78 & 43.69 & 65.69 & 56.35 & 59.55 & 73.37 & 47.38 & 10.57 & 2.41 & 0.66 & 44.51 & 9.95 & 22.82 & 90.19 & 90.65 & 90.41 \\
GLM-4.6V & 64.78 & 69.51 & 42.73 & 66.27 & 58.23 & 67.99 & 75.48 & 42.21 & 9.57 & 1.94 & 0.49 & 43.86 & 9.96 & 22.12 & 88.42 & 90.63 & 89.50 \\
\groupsep
PaddleOCR-VL & 9.56 & 28.25 & 5.24 & 9.04 & 1.60 & 3.75 & 16.63 & 19.05 & 1.82 & 0.05 & 0.01 & 4.47 & 0.47 & 3.66 & 84.52 & 78.70 & 81.48 \\
PaddleOCR-VL-1.5 & 3.73 & 23.11 & 2.52 & 1.03 & 0.33 & 1.44 & 6.21 & 0.34 & 0.00 & 0.00 & 0.00 & 0.22 & 0.00 & 0.22 & 76.81 & 74.76 & 75.68 \\
\groupsep
DeepSeek-VL2-tiny & 51.46 & 71.43 & 29.85 & 37.87 & 42.33 & 49.08 & 71.78 & 47.16 & 10.89 & 2.58 & 0.63 & 38.10 & 8.81 & 20.03 & 90.57 & 89.29 & 89.92 \\
DeepSeek-VL2 & 36.82 & 50.55 & 27.98 & 46.28 & 8.87 & 51.21 & 42.53 & 46.54 & 10.66 & 2.69 & 0.67 & 40.41 & 9.33 & 20.99 & 90.69 & 89.68 & 90.18 \\
DeepSeek-vl-7B & 51.64 & 54.77 & 37.47 & 44.14 & 53.63 & 51.62 & 70.65 & 49.56 & 10.07 & 1.54 & 0.25 & 21.12 & 4.58 & 12.77 & 91.17 & 86.13 & 88.54 \\
DeepSeek-OCR & 4.61 & 24.96 & 2.56 & 1.29 & 1.61 & 0.00 & 12.38 & 40.00 & 8.72 & 1.92 & 0.41 & 35.13 & 7.75 & 19.18 & 89.47 & 88.70 & 89.06 \\
\groupsep
LLaVA-1.5-7b & 44.18 & 49.52 & 40.67 & 33.35 & 56.19 & 42.34 & 55.27 & 53.14 & 13.11 & \second{2.82} & 0.55 & 31.27 & 7.62 & 18.76 & 90.86 & 86.97 & 88.87 \\
LLaVA-NeXT-Video-7B & 14.34 & 30.12 & 6.24 & 7.80 & 0.00 & 14.53 & 34.74 & 35.76 & 7.47 & 1.54 & 0.31 & 37.09 & 7.76 & 19.66 & 89.13 & 89.45 & 89.29 \\
LLaVA-v1.6-mistral-7b & 36.25 & 50.20 & 37.79 & 25.54 & 8.32 & 45.42 & 55.90 & 46.61 & 9.78 & 1.91 & 0.34 & 37.22 & 7.87 & 19.70 & 90.80 & 88.64 & 89.70 \\
LLaVA-OneVision-1.5-8B & 21.53 & 38.36 & 10.24 & 25.33 & 2.24 & 15.71 & 34.74 & 39.63 & 7.87 & 1.79 & 0.40 & 37.61 & 7.52 & 18.79 & 89.12 & 89.39 & 89.25 \\
VideoLLaMA2-7B & 52.47 & 64.97 & 39.50 & 42.94 & 55.44 & 50.56 & 61.80 & 30.19 & 5.70 & 0.77 & 0.12 & 31.64 & 5.98 & 18.04 & 88.61 & 89.09 & 88.85 \\
\groupsep
MiniCPM-V-2-6 & 55.18 & 63.91 & 38.54 & 43.63 & 68.02 & 57.96 & 68.96 & 40.90 & 7.52 & 1.22 & 0.26 & 34.74 & 6.39 & 18.71 & 90.18 & 88.61 & 89.38 \\
MiniCPM-V-4-5 & 26.95 & 40.77 & 18.71 & 29.21 & 5.54 & 29.52 & 40.77 & 49.05 & 12.28 & 3.31 & 0.96 & 43.32 & 10.87 & 22.45 & 90.94 & 90.09 & 90.51 \\
\groupsep
Internlm-xcomposer2-VL-1.8B & 44.51 & 53.30 & 31.08 & 32.20 & 42.97 & 44.13 & 66.96 & \best{60.48} & \best{16.32} & 2.28 & 0.26 & 17.76 & 4.30 & 11.80 & \second{91.75} & 85.15 & 88.31 \\
Internlm-xcomposer2-vl-7B & 21.53 & 36.86 & 10.24 & 25.33 & 2.24 & 15.71 & 34.74 & 13.70 & 0.55 & 0.00 & 0.00 & 11.46 & 0.46 & 8.84 & 82.64 & 83.86 & 83.25 \\
\groupsep
Phi-3.5-Vision & 50.43 & 58.26 & 29.59 & 48.27 & 57.79 & 62.24 & 60.58 & 45.71 & 9.14 & 1.95 & 0.41 & 33.19 & 6.70 & 17.79 & 90.43 & 88.03 & 89.20 \\
Phi3-Vision & 52.37 & 56.35 & 36.09 & 49.94 & 61.73 & 50.79 & 64.20 & 35.81 & 5.48 & 0.29 & 0.00 & 7.47 & 1.05 & 5.87 & 88.12 & 82.05 & 84.96 \\
Phi-4-Multimodal & 60.22 & 69.91 & 45.68 & 55.27 & 63.76 & 57.32 & 67.93 & 51.96 & 11.54 & 2.39 & 0.53 & 36.96 & 8.18 & 19.90 & 91.28 & 88.33 & 89.77 \\
\groupsep
\rowcolor{avgcolor}
\textbf{Model Average} & 52.06 & 61.02 & 36.00 & 50.65 & 44.90 & 52.15 & 64.50 & 41.55 & 9.61 & 2.27 & 0.62 & 35.73 & 8.45 & 18.97 & 88.34 & 87.81 & 88.04 \\
\bottomrule
\end{tabular}%
}
\par\smallskip
\begin{minipage}{\textwidth}
\footnotesize
\raggedright
\textsuperscript{}ILU, Image-Level Understanding; OLR, Object-Level Recognition; GSI, Geomatic \& Spatial Inference; FA, Function Assessment; STE, Spatial-Temporal Evolution; RB, Robustness.
\end{minipage}
\end{table*}

\subsubsection{Perception Assessment}
The Perception capability dimension is designed to evaluate the model’s fundamental ability to interpret images. Table~\ref{tab:perception_results} shows the comprehensive perception results of all 43 evaluated VLMs. This dimension comprises seven leaf tasks, which are further organized into three sub-dimensions: scene-level understanding, object-level recognition, and pixel-level segmentation.

For scene-level understanding, current VLMs already exhibit relatively strong performance. Gemini-3-Flash leads the scene-oriented tasks, reaching 81.66\% on Scene Understanding and 87.85\% on Scene Classification, while GPT-4o and Qwen3-VL-32B-Instruct follow closely on Scene Understanding with 80.81\% and 80.51\%. In contrast, proprietary models no longer maintain a clear advantage on Image Description. Qwen3-VL-235B-Instruct achieves the highest score at 90.89\%, followed by Qwen3-VL-4B with 90.72\%, both exceeding GPT-4o at 90.46\%. In addition, Image Description shows the narrowest performance range among all perception sub-tasks, varying only from 75.68\% to 90.89\%. This result suggests that coarse descriptive ability has become relatively mature, whereas finer-grained scene recognition remains more effective for distinguishing model capability.

\newcommand{\rrsdashline}{%
\cdashline{1-11}[1.5pt/2pt]
}

\begin{table*}[!t]
\centering
\scriptsize
\definecolor{bestcolor}{RGB}{248,215,218}
\definecolor{secondcolor}{RGB}{219,233,251}
\definecolor{avgcolor}{RGB}{226,239,218}
\definecolor{groupcolor}{RGB}{242,242,242}

\caption{Perception-dimension performance comparison of 43 representative VLMs on RRS-10K.}
\label{tab:perception_results}

\resizebox{\textwidth}{!}{%
\begin{tabular}{lccrrrrrrrr}
\toprule
\textbf{Model} & \textbf{Family} & \textbf{Parameters} & \textbf{Avg} & \textbf{SU} & \textbf{SC} & \textbf{ID} & \textbf{OC} & \textbf{VG} & \textbf{AR} & \textbf{OE} \\
\midrule

\rowcolor{groupcolor}
\multicolumn{11}{l}{\textbf{Proprietary Models}} \\
GPT-5.4 & GPT & Prop. & 60.11 & 60.98 & 41.15 & 59.61 & 59.28 & \cellcolor{bestcolor}\textbf{49.89} & 64.18 & \textbf{85.71} \\
GPT-4o & GPT & Prop. & 61.98 & \cellcolor{secondcolor}\textbf{80.81} & 77.19 & 90.46 & 61.41 & 14.93 & 63.54 & 66.31 \\
Claude Sonnet 4.6 & Claude & Prop. & 57.43 & 58.00 & 69.51 & 85.95 & 36.89 & \cellcolor{secondcolor}\textbf{25.80} & 59.06 & 74.84 \\
Gemini-3-Flash & Gemini & Prop. & 64.63 & \cellcolor{bestcolor}\textbf{81.66} & \cellcolor{bestcolor}\textbf{87.85} & 84.84 & 59.28 & 0.00 & \cellcolor{bestcolor}\textbf{71.22} & \cellcolor{secondcolor}\textbf{85.93} \\
Grok-4.1 & Grok & Prop. & 51.50 & 48.61 & 65.88 & 87.81 & 52.45 & 12.58 & 51.17 & 46.06 \\

\rowcolor{groupcolor}
\multicolumn{11}{l}{\textbf{Open-source Models}} \\
Qwen3-VL-2B & Qwen & 2B & 44.41 & 41.58 & 49.68 & 90.35 & 25.37 & 0.00 & 40.30 & 55.01 \\
Qwen3-VL-4B & Qwen & 4B & 50.47 & 49.25 & 55.65 & \cellcolor{secondcolor}\textbf{90.72} & 46.70 & 0.21 & 55.44 & 52.24 \\
Qwen2.5-VL-7B-Instruct & Qwen & 7B & 56.52 & 68.95 & 64.39 & 90.07 & 51.29 & 0.00 & 56.20 & 53.42 \\
Qwen3-VL-8B & Qwen & 8B & 61.65 & 61.83 & 68.02 & 90.57 & \cellcolor{secondcolor}\textbf{62.47} & 0.00 & 59.28 & 60.98 \\
Qwen3-VL-8B-Instruct & Qwen & 8B & 62.56 & 62.96 & 72.71 & 90.59 & 59.87 & 0.00 & 59.40 & 51.28 \\
Qwen3-VL-8B-Thinking & Qwen & 8B & 50.63 & 45.61 & 61.19 & 90.20 & 55.58 & 0.22 & 57.48 & 47.44 \\
Qwen3-VL-30B-A3B-Instruct & Qwen & 30B & 56.00 & 54.82 & 78.25 & 90.24 & 52.79 & 0.00 & 58.97 & 48.48 \\
Qwen3-VL-30B-A3B-Thinking & Qwen & 30B & 55.20 & 64.24 & 67.80 & 88.82 & 55.15 & 0.00 & 64.10 & 45.73 \\
Qwen3-Omni-30B-A3B-Captioner & Qwen & 30B & 52.39 & 56.53 & \cellcolor{secondcolor}\textbf{78.68} & 89.53 & 42.70 & 0.43 & 51.45 & 51.66 \\
Qwen3-Omni-30B-A3B-Instruct & Qwen & 30B & 50.57 & 69.16 & 68.23 & 90.30 & 46.35 & 0.22 & 47.79 & 55.34 \\
Qwen3-Omni-30B-A3B-Thinking & Qwen & 30B & 61.94 & 68.52 & 72.28 & 90.26 & 60.09 & 3.70 & 61.39 & 47.01 \\
Qwen3-VL-32B-Thinking & Qwen & 32B & 55.30 & 65.52 & 62.26 & 90.09 & 58.80 & 0.00 & 61.11 & 51.50 \\
Qwen2.5-VL-32B-Instruct & Qwen & 32B & 54.67 & 70.45 & 59.06 & 87.28 & 55.15 & 0.64 & 56.41 & 55.56 \\
Qwen3-VL-32B-Instruct & Qwen & 32B & 65.07 & 80.51 & 60.55 & 90.36 & 60.73 & 0.00 & \cellcolor{secondcolor}\textbf{68.16} & 62.18 \\
Qwen2-VL-72B-Instruct & Qwen & 72B & 56.28 & 49.04 & 65.97 & 90.06 & 51.29 & 12.24 & 60.09 & 64.06 \\
Qwen2.5-VL-72B-Instruct & Qwen & 72B & \cellcolor{bestcolor}\textbf{65.71} & 79.01 & 61.52 & 90.50 & 62.50 & 7.29 & 61.33 & 58.87 \\
Qwen3-VL-235B-Instruct & Qwen & 235B & \cellcolor{secondcolor}\textbf{65.16} & 70.53 & 64.49 & \cellcolor{bestcolor}\textbf{90.89} & 56.31 & 0.00 & 64.22 & 56.30 \\
Qwen3-VL-235B-Thinking & Qwen & 235B & 57.46 & 68.93 & 67.82 & 90.13 & \cellcolor{bestcolor}\textbf{63.52} & 0.86 & 64.32 & 48.50 \\

\rrsdashline
GLM-4.5V & GLM & 9B & 55.36 & 59.90 & 61.86 & 90.41 & 47.60 & 0.66 & 61.38 & 65.10 \\
GLM-4.6V & GLM & 9B & 55.53 & 66.81 & 56.93 & 89.50 & 43.13 & 0.21 & 61.32 & 66.24 \\

\rrsdashline
PaddleOCR-VL & PaddleOCR & 0.9B & 16.66 & 1.28 & 20.68 & 81.48 & 6.22 & 0.00 & 7.48 & 7.26 \\
PaddleOCR-VL-1.5 & PaddleOCR & 0.9B & 13.62 & 0.86 & 12.15 & 75.68 & 3.90 & 0.00 & 3.42 & 2.74 \\

\rrsdashline
DeepSeek-VL2-tiny & DeepSeek & 3B & 47.54 & 68.87 & 75.48 & 89.92 & 39.23 & 1.28 & 50.11 & 28.78 \\
DeepSeek-VL2 & DeepSeek & 27B & 33.18 & 25.72 & 49.46 & 90.18 & 19.79 & 10.56 & 59.51 & 22.07 \\
DeepSeek-VL-7B & DeepSeek & 7B & 43.03 & 45.20 & 33.69 & 88.54 & 39.02 & 6.18 & 48.40 & 56.29 \\
DeepSeek-OCR & DeepSeek & 3B & 4.53 & 0.21 & 5.97 & 89.06 & 0.64 & 0.00 & 4.27 & 5.34 \\

\rrsdashline
LLaVA-1.5-7B & LLaVA & 7B & 46.02 & 37.10 & 27.93 & 88.87 & 32.41 & 7.25 & 40.72 & 82.30 \\
LLaVA-NeXT-Video-7B & LLaVA & 7B & 17.70 & 0.00 & 16.84 & 89.29 & 10.66 & 0.00 & 0.21 & 14.07 \\
LLaVA-v1.6-Mistral-7B & LLaVA & 7B & 40.48 & 33.69 & 41.15 & 89.70 & 19.83 & 21.32 & 23.24 & \cellcolor{bestcolor}\textbf{86.78} \\
LLaVA-OneVision-1.5-8B & LLaVA & 8B & 28.32 & 1.71 & 40.94 & 89.25 & 10.66 & 0.00 & 15.35 & 14.93 \\
VideoLLaMA2-7B & VideoLLaMA & 7B & 53.05 & 67.38 & 51.17 & 88.85 & 22.17 & 5.54 & 47.97 & 82.30 \\

\rrsdashline
MiniCPM-V-2-6 & MiniCPM & 8B & 50.10 & 69.08 & 42.00 & 89.38 & 43.50 & 2.13 & 55.86 & 52.67 \\
MiniCPM-V-4-5 & MiniCPM & 8B & 22.56 & 4.05 & 41.58 & 90.51 & 20.90 & 7.04 & 27.08 & 19.83 \\

\rrsdashline
InternLM-XComposer2-VL-1.8B & InternLM & 2B & 41.56 & 56.08 & 24.31 & 88.31 & 37.10 & 1.28 & 38.38 & 47.55 \\
InternLM-XComposer2-VL-7B & InternLM & 7B & 22.37 & 1.71 & 40.94 & 83.25 & 10.66 & 0.00 & 15.35 & 14.93 \\

\rrsdashline
Phi-3.5-Vision & Phi & 4B & 42.94 & 49.25 & 44.14 & 89.20 & 26.87 & 2.77 & 46.06 & 42.64 \\
Phi-3-Vision & Phi & 4B & 48.64 & 33.05 & 55.01 & 84.96 & 41.58 & 0.43 & 49.68 & 52.67 \\
Phi-4-Multimodal & Phi & 6B & 57.17 & 71.43 & 58.21 & 89.77 & 46.27 & 5.33 & 56.29 & 74.84 \\

\rowcolor{avgcolor}
\textbf{Model Average} & -- & -- & 47.86 & 50.02 & 53.97 & 88.04 & 40.89 & 4.67 & 48.11 & 50.32 \\
\bottomrule
\end{tabular}%
}

\par\smallskip
\begin{minipage}{\textwidth}
\footnotesize
\raggedright
Prop., proprietary model with undisclosed parameter scale. Family denotes the model series or provider family. SU, Scene Understanding; SC, Scene Classification; ID, Image Description; OC, Object Counting; VG, Visual Grounding; AR, Attribute Recognition; OE, Object Existence.
\end{minipage}
\end{table*}

Object-level recognition remains the main bottleneck within Perception. In Object Counting, the strongest result is obtained by Qwen3-VL-235B-Thinking at 63.52\%, with Qwen2.5-VL-72B-Instruct and Qwen3-VL-8B following closely at 62.50\% and 62.47\%, indicating that counting performance is still moderate even for the best-performing models. Attribute Recognition is led by Gemini-3-Flash with 71.22\%, followed by Qwen3-VL-32B-Instruct at 68.16\%. By comparison, Object Existence is relatively less challenging, with several models exceeding 80\%, including LLaVA-v1.6-mistral-7b at 86.78\%, Gemini-3-Flash at 85.93\%, and GPT-5.4 at 85.71\%. Visual Grounding, however, remains the most difficult perception task. GPT-5.4 reaches only 49.89\%, whereas the second-best Claude Sonnet 4.6 drops to 25.80\%, and GPT-4o further declines to 14.93\%, and the majority of open-source models remain close to zero. Current VLMs are already competent in coarse scene interpretation, but still lack fine-grained grounding capability for rare remote sensing imagery.

For pixel-level segmentation, we further evaluate 9 referring segmentation models on the Referring Remote Sensing Image Segmentation (RRSIS) task, as summarized in Table~\ref{tab:rrsis_results}.

\newsavebox{\rrsistablebox}

\begin{table*}[!t]
\centering
\scriptsize
\renewcommand{\arraystretch}{1.00}
\setlength{\tabcolsep}{3pt}
\caption{Pixel-Level Performance Comparison of 9 Representative Referring Segmentation Models on the Referring Remote Sensing Image Segmentation (RRSIS) Task}
\label{tab:rrsis_results}

\sbox{\rrsistablebox}{%
\resizebox{\textwidth}{!}{%
\begin{tabular}{lccccccccccc}
\toprule
\textbf{Model} & \textbf{Parameters} & \textbf{Pr@50} & \textbf{Pr@60} & \textbf{Pr@70} & \textbf{Pr@80} & \textbf{Pr@90} & \textbf{oIoU} & \textbf{mIoU} & \textbf{Dice} & \textbf{Precision} & \textbf{Recall} \\
\midrule
GeoPixel-Res & 1.3B & \cellcolor{bestcolor}\textbf{37.78} & \cellcolor{bestcolor}\textbf{33.48} & \cellcolor{bestcolor}\textbf{29.64} & \cellcolor{bestcolor}\textbf{22.40} & \cellcolor{secondcolor}\textbf{10.18} & \cellcolor{bestcolor}\textbf{27.31} & \cellcolor{bestcolor}\textbf{35.04} & \cellcolor{bestcolor}\textbf{40.75} & \cellcolor{bestcolor}\textbf{43.38} & 46.66 \\
PaliGemma-448 & 3B & \cellcolor{secondcolor}\textbf{18.55} & \cellcolor{secondcolor}\textbf{15.16} & 12.90 & 8.37 & 1.81 & 6.60 & \cellcolor{secondcolor}\textbf{18.64} & \cellcolor{secondcolor}\textbf{23.31} & \cellcolor{secondcolor}\textbf{20.73} & \cellcolor{secondcolor}\textbf{57.75} \\
PaliGemma-896 & 3B & 16.29 & 14.48 & 11.99 & 8.14 & 2.94 & 5.18 & 17.07 & 21.24 & 18.67 & \cellcolor{bestcolor}\textbf{66.03} \\
PaliGemma-224 & 3B & 14.03 & 11.76 & 8.37 & 4.52 & 0.90 & 5.61 & 15.24 & 19.71 & 16.51 & 57.97 \\
FastSAM & 11.8M & 15.38 & 14.93 & \cellcolor{secondcolor}\textbf{13.80} & \cellcolor{secondcolor}\textbf{13.35} & \cellcolor{bestcolor}\textbf{10.41} & 4.04 & 14.76 & 16.15 & 16.77 & 26.85 \\
CLIPSeg-RD64-Refined & 150M & 8.82 & 4.98 & 2.71 & 0.68 & 0.00 & \cellcolor{secondcolor}\textbf{9.48} & 11.06 & 15.28 & 16.92 & 26.75 \\
CLIPSeg-RD64 & 150M & 6.11 & 4.07 & 1.81 & 0.23 & 0.00 & 6.82 & 9.44 & 13.35 & 14.74 & 31.55 \\
CLIPSeg-RD16 & 150M & 2.04 & 0.68 & 0.68 & 0.23 & 0.00 & 6.04 & 5.39 & 8.34 & 11.18 & 20.60 \\
GLaMM-RefSeg & 7B & 2.49 & 1.36 & 1.13 & 0.90 & 0.00 & 2.59 & 3.73 & 5.43 & 4.37 & 38.25 \\
\rowcolor{avgcolor}
\textbf{Model Average} & -- & 13.50 & 11.21 & 9.23 & 6.54 & 2.92 & 8.19 & 14.49 & 18.17 & 18.14 & 41.38 \\
\bottomrule
\end{tabular}%
}%
}

\usebox{\rrsistablebox}

\par\smallskip
\begin{minipage}{\wd\rrsistablebox}
\footnotesize
\raggedright
Pr@50/60/70/80/90, threshold-based precision scores under different overlap criteria; oIoU, overall intersection over union; mIoU, mean intersection over union; Dice, Dice coefficient. The best and second-best results in each column are highlighted in light red and light blue.
\end{minipage}
\end{table*}

Pixel-level referring segmentation remains highly challenging in rare remote sensing scenes, with substantial performance gaps across models. GeoPixel-Res shows the strongest overall performance and clearly dominates most evaluation metrics, achieving the best results. By contrast, FastSAM performs best only under the strictest overlap criterion, reaching the highest Pr@90, which suggests relatively better high-overlap predictions on a limited subset of samples, but weaker overall mask quality. The PaliGemma series exhibits comparatively high Recall, with PaliGemma-896 reaching the highest Recall of 66.03\%, yet its lower IoU and Precision yet its lower IoU. CLIPSeg-based variants and GLaMM-RefSeg perform substantially worse across most metrics. Overall, the low model-average scores, particularly 8.19 on oIoU and 14.49 on mIoU, indicate that current models still struggle to establish accurate pixel-level alignment for rare remote sensing scenes.

\subsubsection{Reasoning Assessment}

The reasoning dimension presents the comprehensive reasoning results across all evaluated VLMs. The reasoning dimension covers nine leaf tasks spanning three capability groups, including Geometric \& Spatial Inference, Function Assessment, and Spatial-Temporal Evolution. As illustrated in~\ref{tab:reasoning_results}.

\begin{table*}[!t]
\centering
\scriptsize
\renewcommand{\arraystretch}{1.00}
\setlength{\tabcolsep}{3pt}
\definecolor{bestcolor}{RGB}{248,215,218}
\definecolor{secondcolor}{RGB}{219,233,251}
\definecolor{avgcolor}{RGB}{226,239,218}
\definecolor{groupcolor}{RGB}{242,242,242}
\caption{Reasoning-dimension performance comparison of representative VLMs on RRS-10K.}
\label{tab:reasoning_results}
\resizebox{\textwidth}{!}{%
\begin{tabular}{lccrrrrrrrrrr}
\toprule
\textbf{Model} & \textbf{Family} & \textbf{Parameters} & \textbf{Avg} & \textbf{API} & \textbf{RSI} & \textbf{ASI} & \textbf{AC} & \textbf{DC} & \textbf{UI} & \textbf{AI} & \textbf{TAI} & \textbf{CI} \\
\midrule

\rowcolor{groupcolor}
\multicolumn{13}{l}{\textbf{Proprietary Models}} \\
GPT-5.4 & GPT & Prop. & \cellcolor{bestcolor}\textbf{72.64} & \cellcolor{bestcolor}\textbf{50.32} & \cellcolor{bestcolor}\textbf{93.39} & \cellcolor{secondcolor}\textbf{74.84} & 64.04 & 80.60 & 39.66 & 30.28 & 55.44 & \cellcolor{bestcolor}\textbf{94.43} \\
GPT-4o & GPT & Prop. & 59.76 & 32.41 & 83.58 & 67.16 & 69.60 & 80.17 & \cellcolor{secondcolor}\textbf{63.75} & 73.13 & 50.32 & 83.08 \\
Claude Sonnet 4.6 & Claude & Prop. & 62.17 & 38.59 & 88.49 & 70.79 & 62.58 & 72.71 & 40.94 & 49.89 & 55.44 & 86.30 \\
Gemini-3-Flash & Gemini & Prop. & 67.47 & \cellcolor{secondcolor}\textbf{48.83} & 89.13 & 74.41 & \cellcolor{secondcolor}\textbf{71.43} & 75.91 & \cellcolor{bestcolor}\textbf{66.74} & 87.42 & 51.39 & 87.37 \\
Grok-4.1 & Grok & Prop. & 54.05 & 35.61 & 60.77 & 53.09 & 54.27 & 66.52 & 34.97 & 41.36 & 38.38 & 73.66 \\

\rowcolor{groupcolor}
\multicolumn{13}{l}{\textbf{Open-source Models}} \\
Qwen3-VL-2B & Qwen & 2B & 34.50 & 14.50 & 66.74 & 27.29 & 10.45 & 34.97 & 10.02 & 78.04 & 25.80 & 56.53 \\
Qwen3-VL-4B & Qwen & 4B & 57.75 & 40.30 & 81.66 & 63.54 & 64.61 & 72.71 & 26.44 & 89.34 & 38.59 & \cellcolor{secondcolor}\textbf{91.22} \\
Qwen3-VL-8B & Qwen & 8B & 60.15 & 39.45 & 82.52 & 58.21 & 66.95 & 78.04 & 33.26 & 91.90 & 35.39 & 89.29 \\
Qwen3-VL-8B-Instruct & Qwen & 8B & 61.15 & 37.42 & 82.01 & 57.91 & 64.61 & 75.64 & 34.12 & 91.88 & 35.55 & 87.10 \\
Qwen3-VL-8B-Thinking & Qwen & 8B & 58.18 & 43.66 & 84.33 & 62.50 & 57.48 & 76.71 & 35.04 & 69.87 & 44.33 & 66.45 \\
Qwen3-VL-30B-A3B-Instruct & Qwen & 30B & 61.19 & 27.53 & 88.22 & 72.27 & 64.25 & 77.26 & 41.11 & 84.25 & 54.00 & 88.73 \\
Qwen3-VL-30B-A3B-Thinking & Qwen & 30B & 62.47 & 38.71 & 86.51 & 64.96 & 66.10 & 73.50 & 33.97 & 78.21 & 44.75 & 81.72 \\
Qwen3-Omni-30B-A3B-Captioner & Qwen & 30B & 59.36 & 24.83 & 81.58 & 65.47 & 66.96 & 75.72 & 37.53 & 89.02 & 47.01 & 90.03 \\
Qwen3-Omni-30B-A3B-Instruct & Qwen & 30B & 62.07 & 24.01 & 83.60 & 66.30 & 65.36 & 78.07 & 37.34 & 92.14 & \cellcolor{secondcolor}\textbf{56.12} & 90.58 \\
Qwen3-Omni-30B-A3B-Thinking & Qwen & 30B & 64.37 & 38.71 & 86.12 & 66.45 & 66.95 & 80.77 & 43.16 & 74.79 & 49.89 & 82.33 \\
Qwen3-VL-32B-Instruct & Qwen & 32B & 63.91 & 40.00 & 85.65 & 65.17 & 58.64 & 80.98 & 38.25 & 81.84 & 49.25 & 87.74 \\
Qwen3-VL-32B-Thinking & Qwen & 32B & 58.68 & 48.39 & 87.58 & 67.09 & 53.73 & 73.72 & 32.69 & 73.93 & 35.19 & 74.30 \\
Qwen2-VL-72B-Instruct & Qwen & 72B & 60.00 & 45.14 & 86.83 & 65.18 & 69.42 & 80.89 & 41.30 & 55.32 & 36.06 & 81.17 \\
Qwen2.5-VL-32B-Instruct & Qwen & 32B & 58.86 & 26.02 & 85.44 & 64.32 & 67.38 & 77.48 & 44.23 & 70.52 & 38.33 & 88.29 \\
Qwen2.5-VL-72B-Instruct & Qwen & 72B & 66.30 & 37.70 & 89.80 & 73.37 & \cellcolor{bestcolor}\textbf{72.82} & 80.47 & 38.12 & 55.26 & 45.26 & 89.29 \\
Qwen3-VL-235B-Instruct & Qwen & 235B & \cellcolor{secondcolor}\textbf{68.19} & 37.71 & \cellcolor{secondcolor}\textbf{90.95} & \cellcolor{bestcolor}\textbf{77.02} & 70.00 & \cellcolor{bestcolor}\textbf{85.41} & 48.51 & 75.83 & \cellcolor{bestcolor}\textbf{62.60} & 88.99 \\
Qwen3-VL-235B-Thinking & Qwen & 235B & 63.21 & 47.31 & 88.44 & 64.53 & 68.02 & \cellcolor{secondcolor}\textbf{85.34} & 30.29 & 62.84 & 46.47 & 74.19 \\

\cdashline{1-13}[0.6pt/2pt]
GLM-4.5V & GLM & 9B & 59.31 & 38.12 & 89.33 & 71.53 & 54.32 & 75.13 & 31.93 & 80.77 & 43.48 & 75.61 \\
GLM-4.6V & GLM & 9B & 59.41 & 44.30 & 88.65 & 71.58 & 50.53 & 76.28 & 34.83 & 81.62 & 48.39 & 87.59 \\

\cdashline{1-13}[0.6pt/2pt]
Phi-4-Multimodal & Phi & 6B & 50.69 & 24.31 & 75.48 & 52.67 & 58.64 & 65.25 & 32.41 & \cellcolor{secondcolor}\textbf{95.10} & 34.97 & 79.66 \\
Phi3-Vision & Phi & 4B & 47.71 & 25.80 & 68.66 & 34.75 & 48.61 & 71.86 & 26.65 & \cellcolor{bestcolor}\textbf{96.80} & 29.85 & 71.73 \\
Phi-3.5-Vision & Phi & 4B & 43.25 & 29.21 & 64.61 & 39.87 & 38.81 & 68.87 & 20.90 & 94.67 & 35.61 & 88.87 \\

\cdashline{1-13}[0.6pt/2pt]
MiniCPM-V-2-6 & MiniCPM & 8B & 44.36 & 31.34 & 51.39 & 21.54 & 42.86 & 71.00 & 39.87 & 96.16 & 33.69 & 82.23 \\
MiniCPM-V-4-5 & MiniCPM & 8B & 19.88 & 18.55 & 52.67 & 42.22 & 1.07 & 31.56 & 4.26 & 6.82 & 14.71 & 44.33 \\

\cdashline{1-13}[0.6pt/2pt]
DeepSeek-vl-7B & DeepSeek & 7B & 39.09 & 28.14 & 58.00 & 49.04 & 26.23 & 59.28 & 35.82 & 71.43 & 44.78 & 58.46 \\
DeepSeek-VL2-tiny & DeepSeek & 3B & 32.86 & 27.29 & 61.62 & 49.68 & 3.20 & 47.55 & 19.40 & 65.25 & 31.13 & 67.02 \\
DeepSeek-VL2 & DeepSeek & 27B & 37.87 & 26.60 & 76.34 & 57.61 & 24.21 & 46.62 & 6.05 & 11.68 & 39.73 & 62.68 \\
DeepSeek-OCR & DeepSeek & 3B & 1.19 & 2.15 & 0.43 & 0.00 & 0.43 & 3.42 & 2.14 & 1.07 & 0.00 & 0.00 \\

\cdashline{1-13}[0.6pt/2pt]
Internlm-xcomposer2-VL-1.8B & InternLM & 2B & 31.44 & 20.90 & 50.11 & 26.87 & 27.29 & 35.82 & 15.78 & 70.15 & 40.51 & 47.75 \\
Internlm-xcomposer2-vl-7B & InternLM & 7B & 18.65 & 18.12 & 51.17 & 55.44 & 0.00 & 1.92 & 4.05 & 0.43 & 12.79 & 18.63 \\

\cdashline{1-13}[0.6pt/2pt]
LLaVA-v1.6-mistral-7b & LLaVA & 7B & 26.40 & 18.12 & 52.67 & 47.33 & 0.00 & 9.59 & 12.15 & 4.48 & 40.09 & 50.75 \\
LLaVA-1.5-7b & LLaVA & 7B & 37.07 & 28.57 & 62.47 & 49.47 & 2.13 & 24.09 & 23.67 & 88.70 & 28.57 & 56.10 \\
LLaVA-OneVision-1.5-8B & LLaVA & 8B & 18.65 & 18.12 & 51.17 & 55.44 & 0.00 & 1.92 & 4.05 & 0.43 & 12.79 & 18.63 \\
LLaVA-NeXT-Video-7B & LLaVA & 7B & 10.82 & 11.51 & 6.40 & 20.68 & 0.00 & 0.43 & 0.00 & 0.00 & 13.43 & 15.63 \\
VideoLLaMA2-7B & VideoLLaMA & 7B & 45.28 & 31.13 & 67.59 & 31.98 & 30.70 & 53.30 & 29.85 & 81.02 & 37.74 & 63.38 \\

\cdashline{1-13}[0.6pt/2pt]
PaddleOCR-VL & PaddleOCR & 0.9B & 6.24 & 2.37 & 31.91 & 10.47 & 0.00 & 0.43 & 2.99 & 0.21 & 7.49 & 0.00 \\
PaddleOCR-VL-1.5 & PaddleOCR & 0.9B & 1.19 & 0.00 & 4.71 & 0.22 & 0.00 & 0.21 & 0.43 & 0.23 & 2.88 & 0.00 \\

\rowcolor{avgcolor}
\textbf{Model Average} & -- & -- & 47.55 & 31.90 & 71.84 & 54.39 & 45.02 & 59.76 & 28.43 & 62.75 & 37.73 & 67.70 \\
\bottomrule
\end{tabular}%
}
\par\smallskip
\begin{minipage}{\textwidth}
\footnotesize
\raggedright
Prop., proprietary model with undisclosed parameter scale. Family denotes the model series or provider family. API, Absolute Position Inference; RSI, Relative Spatial Inference; ASI, Absolute Spatial Inference; AC, Area Calculation; DC, Distance Calculation; UI, Usage Inference; AI, Attribute Inference; TAI, Temporal Attribute Inference; CI, Change Identification.
\end{minipage}
\end{table*}

Geometric-spatial reasoning tasks reflect a clear stratification among current models. Relative Spatial Inference (RSI) is the most mature sub-task, with GPT-5.4 achieving 93.39\%, followed by Qwen3-VL-235B-Instruct at 90.95\% and Qwen2.5-VL-72B-Instruct at 89.80\%. Distance Calculation (DC) also shows relatively strong performance for frontier models, with Qwen3-VL-235B-Instruct and Qwen3-VL-235B-Thinking reaching 85.41\% and 85.34\%, respectively, slightly surpassing GPT-5.4. By contrast, Absolute Position Inference (API) and Absolute Spatial Inference (ASI) remain clearly more difficult, indicating that current models still face challenges in precise geospatial reasoning. Area Calculation (AC) further exposes the instability, although Qwen2.5-VL-72B-Instruct reaches 72.82\% and Gemini-3-Flash obtains 71.43\%, performance drops sharply for smaller models, suggesting that quantitative spatial reasoning remains a major bottleneck.

Function-oriented reasoning shows a clear internal imbalance across the evaluated models. Usage Inference (UI) is led by Gemini-3-Flash at 66.74\%, followed by GPT-4o at 63.75\%, while most open-source models remain substantially lower on this task. In contrast, Attribute Inference (AI) is much stronger overall and appears easier for a broad range of models. Phi3-Vision achieves the highest of 96.80\%, followed closely by MiniCPM-V-2-6 at 96.16\% and Phi-4-Multimodal at 95.10\%; several Qwen-family models also exceed 90\%. This large gap between UI and AI suggests that current VLMs are better at inferring static visual or semantic attributes than at reasoning about functional utilization.

In temporal reasoning tasks, model performance also shows a clear difference in difficulty. In Change Identification (CI), GPT-5.4 achieves the best result of 94.43\%, while Qwen3-VL-4B reaches 91.22\% and Qwen3-Omni-30B-A3B-Instruct obtains 90.58\%, indicating that temporal change detection has become relatively tractable for several strong models. However, Temporal Attribute Inference (TAI) is considerably more challenging. Qwen3-VL-235B-Instruct attains the highest score of 62.60\%, followed by Qwen3-Omni-30B-A3B-Instruct at 56.12\%, whereas most other models remain below 50\%. This indicates that recognizing whether change exists is notably easier than inferring deeper temporal properties from rare remote sensing imagery.

\subsubsection{Robustness Assessment}

The robustness dimension presents the comprehensive robustness results across all evaluated VLMs. This dimension covers three sub-tasks, including Noise Robustness (NR), Occlusion Robustness (OR), and Hallucination Detection (HD).

\begin{figure*}[!h]
\includegraphics[width=\textwidth]{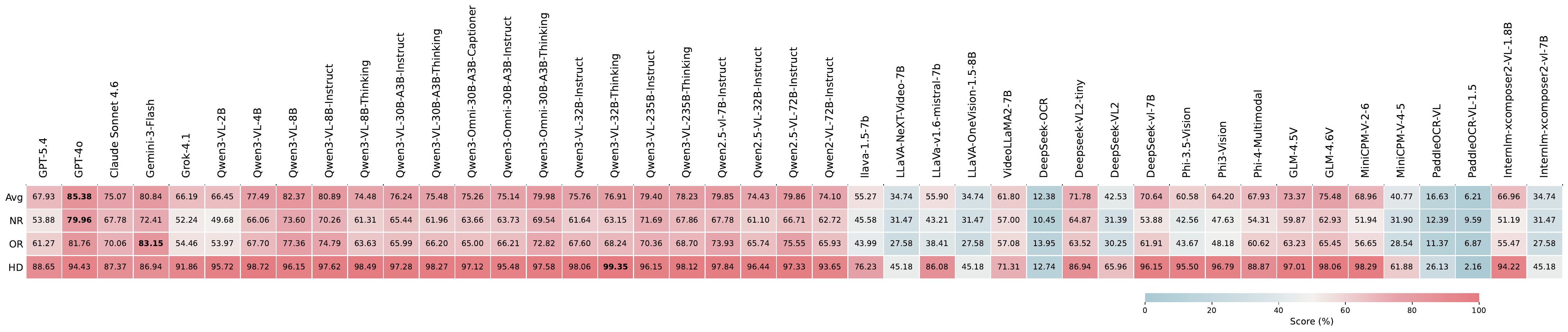}
\caption{Robustness dimension performance heatmap across 43 VLMs. Models are arranged as columns (sorted by average score), with four metrics as rows: Average, Noise Robustness (NR), Occlusion Robustness (OR), and Hallucination Detection (HD).\label{fig9}}
\end{figure*}

As illustrated in Fig.~\ref{fig9}, noise robustness remains a major bottleneck under degraded conditions. GPT-4o achieves the best NR score of 79.96\%, while the strongest competing open-source models remain only in the low-70\% range. Although Gemini-3-Flash and several Qwen models maintain competitive performance, most models show a clear drop, indicating that rare-scene understanding remains highly sensitive to perturbations that blur structural cues.

In the occlusion robustness task, GPT-4o achieves the highest OR score of 81.76\%, followed by Gemini-3-Flash at 80.15\%, while only a limited number of open-source models remain above 75\%. For many other models, OR falls into the 40\%--60\% range or lower, suggesting that they still rely heavily on local visual evidence, easily destabilized when key contextual cues are partially masked. Taken together with the NR results, this indicates that robustness to incomplete observations remains a major unresolved challenge.

Hallucination detection is the clearest strength among the three robustness sub-tasks. Several Qwen-family models achieve HD scores above 98\%, with the best reaching 99.35\%, substantially exceeding GPT-4o at 94.43\% and GPT-5.4 at 88.65\%, indicating that models are effective at rejecting nonexistent targets in deceptive descriptions. The large gap between HD and NR/OR indicates that hallucination suppression is easier than robustness to image degradation or occlusion.
\section{Discussion}
\label{sec:Discussion}

\subsection{Open-Source Models Are Strong Baselines for Rare Remote Sensing}
The results indicate a reduced performance gap between proprietary models and leading open-source models on rare remote sensing interpretation. In particular, several leading open-source models, especially the Qwen series, achieve performance comparable to proprietary systems, even surpassing them on typical tasks. This finding suggests that high-performing open-source models can therefore serve as practical baselines for future studies on RS-VLMs. Additionally, model scale is still positively related to overall performance, but parameter count alone cannot explain the observed differences. Larger models usually achieve better average results, yet the gains are inconsistent across tasks, and some mid-scale models remain more parameter-efficient than much larger counterparts. This indicates that architecture, multimodal alignment, and instruction tuning are also decisive factors. Future progress in rare-scene remote sensing should depend on both effective scaling and better model design.

\subsection{Complex Semantic Understanding and Robustness Remain Major Bottlenecks}
Although several models excel on coarse-grained perception tasks, performance drops markedly on demanding tasks. The largest weaknesses appear in visual grounding, referring segmentation, and higher-order reasoning over spatial relations. The gap is mainly attributable to the difficulty of aligning visual evidence with precise semantic queries in complex rare scenes. 
Robustness is another major limitation. Hallucination detection achieves relatively high scores across several strong models, whereas noise and occlusion robustness remain poor. The result suggests that suppressing deceptive predictions is easier than preserving reliable recognition when visual evidence is degraded. In practice, current models are still highly sensitive to corrupted cues. Rare-scene understanding, therefore, remains insufficiently robust for real-world use.

\subsection{Diagnostic Value for Future RS-VLM Research}
Recent benchmark studies increasingly suggest that the value of evaluation lies not only in ranking models, but also in diagnosing concrete capability bottlenecks and informing model refinement. In this sense, RRS-10K should be viewed as a diagnostic benchmark for rare-scene RS-VLMs' research rather than a static test set for score comparison. Our results show that current models remain relatively reliable on tasks that rely on coarse global scene context, yet degrade sharply on visual grounding, referring segmentation, and structured spatiotemporal reasoning. This pattern indicates that the main limitation of current RS-VLMs in rare scenes is the weak coupling between global scene semantics and fine-grained local evidence. Therefore, RRS-10K provides an empirical basis for future work on the construction of rare-scene data.
\section{Conclusion}
\label{sec:Conclusion}
In this article, we propose RRS-10K, a comprehensive benchmark for rare remote sensing scene interpretation. It contains 10,738 carefully annotated images and provides a hierarchical evaluation framework spanning perception, reasoning, and robustness. To enhance benchmark reliability, a human-centered hybrid annotation pipeline and SDFS were adopted for high-quality MCQ design. Extensive experiments on 43 VLMs and 9 referring segmentation models demonstrate that rare remote sensing scenes remain challenging for current models, especially in complex semantic tasks. The results also indicate that leading open-source models are becoming competitive with proprietary systems, while model scale alone is insufficient for robust rare-scene understanding. Overall, RRS-10K offers a standardized benchmark for future RS-VLM research. Future work will incorporate broader sensing modalities, richer temporal settings, and more diverse real-world rare-scene applications.

\section*{Acknowledgment}
The authors gratefully acknowledge 
Hefei University of Technology and the Institute of Artificial Intelligence, Hefei Comprehensive National Science Center, for support in providing computational resources. 
The authors would also like to thank He Wang and Jingbo Chen from the National University of Defense Technology for their valuable suggestions that helped refine the conceptual development of this work.




\bibliographystyle{IEEEtran}
\bibliography{IEEEtrans}


\begin{IEEEbiographynophoto}{Yuqiao Lai}
 received the B.S. degree in military science from the National University of Defense Technology, Nanjing, China, in 2025, where he is currently pursuing the M.S. degree in military science. His research interests include computer vision, machine learning, and applications of vision-language models to remote sensing.
\end{IEEEbiographynophoto}

\begin{IEEEbiographynophoto}{Jiancheng Qi}
received the B.S. and M.S. degrees in engineering from Information Engineering University, China, in 2006 and 2009, respectively. He is currently pursuing the Ph.D. degree in military science. He is an Associate Professor  with the National University of Defense Technology, Nanjing, China. His research interests include computer vision and remote sensing image interpretation.
\end{IEEEbiographynophoto}

\begin{IEEEbiographynophoto}{Fei Wang}
is currently pursuing the Ph.D. degree in engineering with the School of Computer Science and Information Engineering, Hefei University of Technology, Hefei, China. His research interests include computer vision and multimodal affective computing. He has published six papers at top international conferences, including CVPR, AAAI, IJCAI, and WWW, and has received seven competition awards from ACM MM and IJCAI. He regularly serves as a Program Committee Member for top-tier conferences in multimedia and artificial intelligence, such as ICLR, IJCAI, and ACM MM.
\end{IEEEbiographynophoto}

\begin{IEEEbiographynophoto}{Yuxin Liu}
is with the Institute of Artificial Intelligence, Hefei Comprehensive National Science Center, Hefei, China. His research interests include large language models, multimodal foundation models, multi-agent systems, psychological counseling support systems, and deepfake detection. He has participated in research on large-model fine-tuning and inference deployment, multi-stage agent architecture design, structured memory mechanisms, and evaluation framework construction for complex real-world applications. His related research has been published in ACM WWW 2026 and IEEE CVPR 2026.
\end{IEEEbiographynophoto}

\begin{IEEEbiographynophoto}{Kun Li}
(Member, IEEE) is currently a Postdoctoral Fellow at the United Arab Emirates University. He received the Ph.D. degree from Hefei University of Technology in 2023. His research interests include multimedia content analysis, human-centric video understanding, and affective computing. He regularly serves as a Program Committee Member for top-tier conferences in computer vision and multimedia, such as CVPR, ICCV, ECCV, ICLR, NeurIPS, and ACM Multimedia.
\end{IEEEbiographynophoto}

\begin{IEEEbiographynophoto}{Ye Chen}
received the B.L. degree in law in 2004 and the M.A. degree in literature in 2007, both from the National University of Defense Technology, China. She is currently an Associate Professor with the National University of Defense Technology, Nanjing, China. Her research interests include the fundamental theory of informatics.
\end{IEEEbiographynophoto}

\begin{IEEEbiographynophoto}{Yan Gao}
received the B.S. degree in management science and engineering from Harbin Engineering University, Harbin, China, in 2013, and the Ph.D. degree in management science and engineering from Harbin Engineering University in 2020. She is currently an Associate Professor with the National University of Defense Technology, Nanjing, China. Her research interests include multi-source information fusion and information processing.
\end{IEEEbiographynophoto}

\begin{IEEEbiographynophoto}{Yanyan Wei}
is a Lecturer and Master's Supervisor with the School of Computer Science and Information Engineering, Hefei University of Technology, Hefei, China. He received the Ph.D. degree in engineering in 2022. His research interests include robust image perception and understanding, large models and multi-agent applications, and AI-enabled interdisciplinary applications. He has published more than 30 papers in top journals and conferences, including IEEE TIP, IEEE TMM, ACM TOMM, PR, CVIU, AAAI, WWW, and ACM MM, and has received six competition awards from CVPR, IJCAI, and ACM MM. He serves as a reviewer for leading journals and conferences, including IJCV, CVPR, ICCV, ICLR, and AAAI. He is a member of ACM, CCF, CSIG, and CAAI.
\end{IEEEbiographynophoto}

\end{document}